\documentclass[submission]{article}
\usepackage{ijcai25}

\usepackage{times}
\usepackage{soul}
\usepackage{url}
\usepackage[utf8]{inputenc}
\usepackage[small]{caption}
\usepackage{graphicx}
\usepackage{amsmath}
\usepackage{amsthm}
\usepackage{algorithm}
\usepackage{times}
\usepackage{soul}
\usepackage{url}
\usepackage[hidelinks]{hyperref}
\hypersetup{colorlinks=true,
citecolor=green,
linkcolor=red,
urlcolor=blue
}

\usepackage{graphicx}
\usepackage{makecell}
\usepackage{array}
\usepackage{adjustbox} % For better alignment control

\usepackage{amsmath,amsfonts}
\usepackage{algpseudocode}
\usepackage{bm}
\usepackage{float}
\usepackage{multirow}
\usepackage{makecell}
\usepackage{tabularx}
\usepackage{amssymb}
\usepackage{stmaryrd}
\usepackage{wrapfig}
\usepackage{lipsum}
\usepackage{subfigure}
\usepackage{bbold}
\usepackage{amsthm}
\usepackage{newfloat}
\usepackage{listings}
\usepackage{tikz}
\usepackage{graphicx}
\usepackage{mathtools}
\usepackage{longtable}
\usepackage[most]{tcolorbox}
\usepackage{booktabs} % 优化表格样式
\usepackage{array} % 额外表格控制
\usepackage{subcaption}

\title{Learn to Think: Bootstrapping LLM Reasoning Capability Through Graph Representation Learning}

\author{
Hang Gao$^{1,2}$\thanks{Equal contribution.}\and
Chenhao Zhang$^{1,2,3}$\footnotemark[1]\and
Tie Wang$^4$\thanks{Corresponding authors.}\and
Junsuo Zhao$^{1,2,3}$\and
Fengge Wu$^{1,2,3}$\footnotemark[2]\and
Changwen Zheng$^{1,2,3}$\And
Huaping Liu$^5$\\
\affiliations
$^1$ Institute of Software, Chinese Academy of Sciences.\\
$^2$ National Key Laboratory of Space Integrated Information System.\\
$^3$ University of Chinese Academy of Sciences.\\
$^4$ Peking University.\\
$^5$ Tsinghua University. \\
\emails
\{gaohang, zhangchenhao2024, fengge, changwen, junsuo\}@iscas.ac.cn,\\
wangtie2021@stu.pku.edu.cn,
hpliu@tsinghua.edu.cn
}

\begin{document}

\maketitle

\begin{abstract}
Large Language Models (LLMs) have achieved remarkable success across various domains. However, they still face significant challenges, including high computational costs for training and limitations in solving complex reasoning problems. Although existing methods have extended the reasoning capabilities of LLMs through structured paradigms, these approaches often rely on task-specific prompts and predefined reasoning processes, which constrain their flexibility and generalizability. To address these limitations, we propose a novel framework that leverages graph learning to enable more flexible and adaptive reasoning capabilities for LLMs. Specifically, this approach models the reasoning process of a problem as a graph and employs LLM-based graph learning to guide the adaptive generation of each reasoning step. To further enhance the adaptability of the model, we introduce a Graph Neural Network (GNN) module to perform representation learning on the generated reasoning process, enabling real-time adjustments to both the model and the prompt. Experimental results demonstrate that this method significantly improves reasoning performance across multiple tasks without requiring additional training or task-specific prompt design. \textit{Code can be found in \url{https://github.com/zch65458525/L2T}.}
\end{abstract}

\section{Introduction}

In recent years, LLMs \cite{radford2018improving} have achieved remarkable success in fields such as natural language processing \cite{brown2022language}, machine translation \cite{jiao2022machine}, and code generation \cite{ni2022code}. However, training these models requires substantial computational resources and energy, resulting in high costs and environmental impacts \cite{patterson2022carbon}. As a result, efficiently utilizing LLMs has become a key research focus, with prompt engineering emerging as a critical technique \cite{liu2023pre,zhou2022learning,sun2022black}. By designing effective prompts, it is possible to optimize model performance without additional training, making it a cost-effective and straightforward approach. Notably, the Chain-of-Thought (CoT) method \cite{DBLP:conf/nips/Wei0SBIXCLZ22} has demonstrated significant improvements in tasks such as mathematical reasoning and logical inference by guiding models through step-by-step reasoning processes. CoT works by crafting prompts that break down complex problems into logical steps, enabling the model to solve them incrementally.

% \begin{figure}
%     \centering
%     \includegraphics[width=0.95\linewidth]{motiv1.pdf}
%     \caption{Template}
%     \label{fig:enter-label}
% \end{figure}

% \begin{figure}[htbp]
%     \centering
%     \begin{subfigure}
%         \centering
%         \includegraphics[width=0.2\linewidth]{motiv1.pdf}
%         \caption{Caption for Motiv1}
%         \label{fig:motiv1}
%     \end{subfigure}
%     \begin{subfigure}
%         \centering
%         \includegraphics[width=0.2\linewidth]{motiv2.pdf}
%         \caption{Caption for Motiv2}
%         \label{fig:motiv2}
%     \end{subfigure}
%     \caption{Overall caption for the two images}
%     \label{fig:motiv-combined}
% \end{figure}

\begin{figure}[htbp]
    \centering
    \subfigure[Conventional Method.]{
        \includegraphics[width=0.45\linewidth]{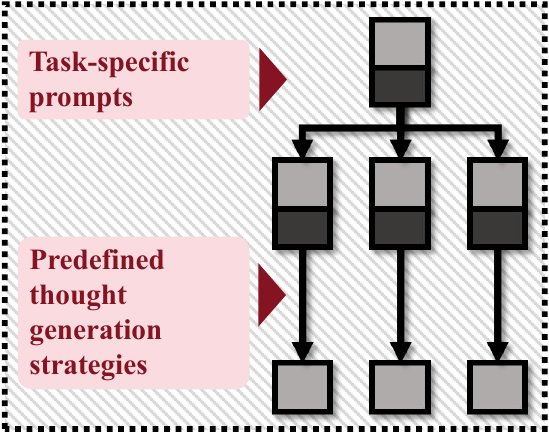}
        \label{fig:motiva} 
    }
    \subfigure[Our Proposed Method.]{
        \includegraphics[width=0.45\linewidth]{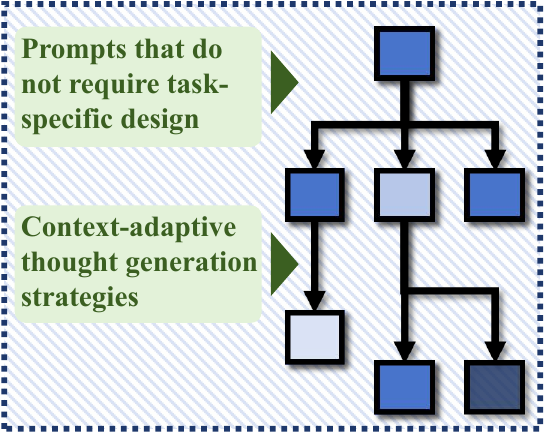}
        \label{fig:motiv2} 
    }
    \vskip -0.1in
    \caption{A comparison between our method and conventional methods.}
    \vskip -0.1in
    \label{fig:motiv}
\end{figure}

Based on Chain of Thoughts, numerous related methods have been proposed in recent years, including Tree of Thoughts (ToT)\cite{DBLP:conf/acl/ChuCCYH0P00L24}, Graph of Thoughts (GoT)\cite{DBLP:conf/aaai/BestaBKGPGGLNNH24}, and Thread of Thoughts (ThoT)\cite{DBLP:journals/corr/abs-2311-08734}. These methods introduce more complex thinking paradigms, such as tree structures, graph structures, and thread-based reasoning, thereby further extending the reasoning capabilities of LLMs. Compared to chain-based reasoning structures, these approaches have significantly enhanced the breadth and depth of the cognition of LLMs \cite{DBLP:conf/acl/QiaoO0CYDTHC23}. They have played an active role in optimizing the performance of LLMs \cite{hadi2024large}. However, these methods still face several critical challenges.

First, they lack adaptability to different contexts. Existing approaches are often unable to make real-time adjustments to models and prompts in response to dynamic changes in scenarios, resulting in limited flexibility and robustness when addressing diverse tasks \cite{DBLP:conf/acl/ChuCCYH0P00L24}. Once the reasoning process begins, LLMs typically follow a predefined prompt and execute reasoning in a relatively fixed manner. This leads to a second issue: these methods often require task-specific prompt design to handle different tasks effectively, particularly for those involving more complex reasoning processes. This reliance is, to some extent, inevitable, as more intricate reasoning demands highly precise and specific prompts to effectively guide the model. Only with carefully crafted prompts can the models fully exploit their extended reasoning frameworks. Without sufficiently targeted prompts, their reasoning performance may degrade greatly, failing to achieve the desired cognitive outcomes. This heavy dependence on task-specific prompts poses a major limitation, severely undermining the generalizability of such methods. One possible solution is to collect task-specific data and use fine-tuning methods to train the LLM. However, this approach incurs significant costs in industrial scenarios and is not feasible for cases where only API access is available. Figure \ref{fig:motiva} provides a visual summary of these challenges.

Thus, a key question emerges: \textbf{is there a way to address different types of problems in a unified manner without requiring LLM training or additional prompt design, while also allowing the model to flexibly adjust based on the problem and reasoning process?} To achieve this, it is essential to establish a suitable unified framework to model the entire reasoning process of LLMs, enabling them to adopt different modes of thinking at appropriate moments, much like humans do.

To this end, we propose the \textit{\textbf{L}earn \textbf{to} \textbf{T}hink} (L2T) method, which guides the LLM to ``think'' based on graph learning. This method employs graphs to unify the representation of the reasoning process of LLMs across different tasks. These graphs are annotatable, enabling more effective representation and accurate prediction of reasoning strategies. 
% The format and evaluation criteria for all contents of the reasoning process are automatically generated by the LLM. 
Subsequently, L2T utilizes a graph learning approach based on LLMs to adaptively guide reasoning strategies for various scenarios. By combining such an approach with the automatic extraction of reasoning process formats and evaluation criteria from task descriptions, L2T effectively handles diverse tasks without relying on task-specific prompts. Then, L2T introduces a GNN-based reasoning mode selection module to perform relatively lightweight representation learning on the graph, facilitating the switch between different reasoning modes for LLMs. This enables real-time adjustments during the reasoning process, and the GNN-based reasoning mode selection module is further refined within a reinforcement learning framework. Figure \ref{fig:motiv2} illustrates the advantages of the proposed method. In summary, our contributions are as follows:

\begin{itemize} 
\item We propose an LLM reasoning framework that can adapt to different problems and develop reasoning pathways without requiring task-specific prompts.
\item By integrating a GNN-based reasoning mode selection module, we enable real-time adjustment of the LLM reasoning strategies. Furthermore, the module can be continuously optimized through reinforcement learning. 
\item Extensive experiments are conducted to thoroughly validate and analyze the proposed method. 
\end{itemize}
% 图X给出了详细解释，同时实验XX 

% 为了解决这个问题，提出基于图表示学习来进行优化 

% “直觉性”的东西只能使用神经网络来建模 
% 图可以覆盖所有类型逻辑
% 任务非特定的
% 可以对于专用任务特化
% 不需要微调大模型

\section{Related Works}
\paragraph{Prompt engineering.}

Prompt engineering for LLMs has seen significant advancements, introducing innovative techniques aimed at enhancing reasoning and reliability. Methods such as CoT \cite{DBLP:conf/nips/Wei0SBIXCLZ22} improve reasoning capabilities by incorporating intermediate steps, while self-consistency \cite{DBLP:conf/iclr/0002WSLCNCZ23} enhances reliability by aggregating consistent outputs. Interactive question answering further enables dynamic interactions with the model, facilitating adaptive reasoning processes \cite{DBLP:conf/iclr/YaoZYDSN023,DBLP:conf/chi/MassonMC024}. To mitigate hallucinations, Retrieval-Augmented Generation (RAG) \cite{DBLP:conf/nips/LewisPPPKGKLYR020} integrates external retrieval mechanisms to ensure factual accuracy. Additionally, methods like Chain-of-Verification (CoVe) \cite{DBLP:conf/acl/DhuliawalaKXRLC24}, Chain-of-Note (CoN) \cite{DBLP:journals/corr/abs-2311-09210}, and Chain-of-Knowledge (CoK) focus on step-by-step validation for robust reasoning. Furthermore, prompt engineering research has also explored areas such as user intent understanding \cite{DBLP:conf/acl/DiaoWLPL024}, autonomous prompt selection \cite{DBLP:conf/iclr/ZhouMHPPCB23}, external tool integration \cite{DBLP:journals/corr/abs-2303-09014}, and emotional control in responses \cite{li2023largelanguagemodelsunderstand}.

\paragraph{Logic and reasoning within LLM prompting.}
Efforts to enhance logic and reasoning in LLM prompting have introduced various innovative methods. Auto-CoT \cite{DBLP:conf/iclr/0001Z0S23} automates the generation of reasoning chains, while Logical CoT (LogiCoT) \cite{DBLP:conf/coling/ZhaoLLW0CW24} leverages symbolic logic for step-by-step verification. Prompt Sketching \cite{DBLP:conf/icml/Beurer-KellnerM24} constrains outputs to predefined logical structures, ensuring coherence and adherence to logical frameworks. Topological frameworks have also been explored, such as ToT \cite{DBLP:conf/nips/YaoYZS00N23} and GoT \cite{DBLP:conf/aaai/BestaBKGPGGLNNH24}, which utilize hierarchical and graph-based structures, respectively, to model complex reasoning processes. Algorithm of Thoughts (AoT) \cite{DBLP:conf/icml/SelAK0024} employs in-context algorithmic examples to guide LLMs through structured reasoning pathways, while ThoT \cite{DBLP:journals/corr/abs-2311-08734} generates structured thought threads to decompose and address complex problems. Although these methods have made significant contributions, they typically follow predefined reasoning processes and depend heavily on task-specific prompts, limiting their adaptability and generalizability. In contrast, our method addresses these limitations by enabling more flexible and adaptive reasoning capabilities for LLMs. 
% Additional details on related work are available in \textbf{Appendix  \ref{apx:erw}}.

% \section{Background}
% \subsection{Graph Representation Learning}

% \subsection{Actor-Critic Method in Reinforcement Learning}

\section{Method}

\begin{figure*}
    \centering
    \includegraphics[width=0.95\linewidth]{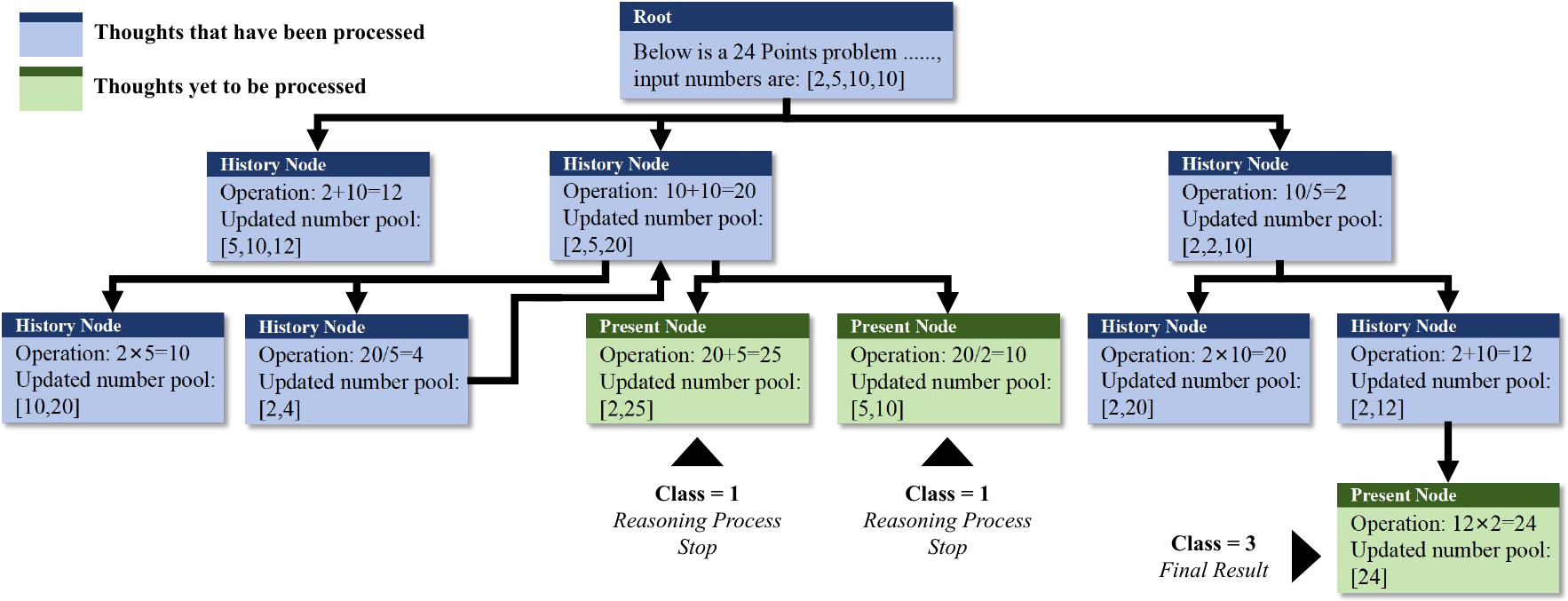}
    \vskip -0.1in
    \caption{An example of the reasoning process graph. Each box contains a thought generated by the LLM, representing a node in the reasoning process graph. The green boxes in the graph indicate the nodes currently being processed. We classify these nodes and used their categories to guide the LLM's next steps.}
    \vskip -0.15in
    \label{fig:logic-graph}
\end{figure*}

% 我们的方法包括三方面内容：1）将LLM的整个逻辑推理过程表示为有向图，2）使用融合图学习的框架对于该有向图进行处理，并灵活、自适应的完成推理，3）基于强化学习对于所提出的推理模型进行不断更新
Our method consists of the following parts: first, representing the complete logical reasoning process of the LLM as a specifically designed graph. Second, automatically generate the format and evaluation criteria of the reasoning process, then employ a graph learning framework to process the reasoning process graph, thereby facilitating flexible and adaptive multi-step problem-solving that does not require task-specific prompts. Finally, iteratively refining the proposed reasoning model through reinforcement learning. We will elaborate on them in detail.

\subsection{Reasoning Process Graph}

\label{sec:reasoning process graph}

The conversation with the LLM consists of user messages (prompts) and the LLM's responses (thoughts). Extensive research has been conducted on how to organize such prompts and thoughts to optimize LLM performance \cite{liu2023pre}, leading to the proposal of various structures of thoughts, such as chain structures \cite{DBLP:conf/nips/Wei0SBIXCLZ22}, tree structures \cite{DBLP:conf/nips/YaoYZS00N23}, graph structures \cite{DBLP:conf/aaai/BestaBKGPGGLNNH24}, etc. Among these, graphs are particularly effective for representing the reasoning frameworks of most existing models, as trees, chains, and other structures can be viewed as special cases of graphs. Building on this, our approach employs a specifically designed graph to represent logical reasoning, which we refer to as the \textit{reasoning process graph}.

Particularly, we represent the entire reasoning process of an LLM as a reasoning process graph \( G = \{\mathcal{V}, \mathcal{E}\} \), where \( \mathcal{V} \) denotes the set of nodes, with each node \( v \in \mathcal{V} \) representing a thought generated by the LLM. Similarly, \( \mathcal{E} \) denotes the set of edges, where each edge \( e \in \mathcal{E} \) represents a connection from one thought to its subsequent thought.

The set \( \mathcal{V} \) can be partitioned into two subsets: \( \mathcal{V}^{\text{pres}} \) and \( \mathcal{V}^{\text{hist}} \). Here, \( \mathcal{V}^{\text{pres}} \) represents the nodes corresponding to unprocessed thoughts and will serve as the basis for generating subsequent thoughts. In contrast, \( \mathcal{V}^{\text{hist}} \) represents the nodes that have already been processed and will no longer be revisited.

Each node \( v \) in \( \mathcal{V}^{\text{pres}} \) is assigned a category label \( Y_v \), where \( Y_v \in \{1, 2, 3, 4\} \). The meaning of each label is as follows:
\begin{itemize}
    \item \textbf{Label 1:} Reasoning should not proceed based on node \( v \).
    \item \textbf{Label 2:} Reasoning should continue based on node \( v \).
    \item \textbf{Label 3:} Node \( v \) should be output as the final result.
    \item \textbf{Label 4:} A backtracking operation should be performed on node \( v \), meaning that reasoning should continue based on its parent node.
\end{itemize}
To assign specific labels to each node in \( \mathcal{V}^{\text{pres}} \), we utilize LLM-based graph learning for node classification. These labels are subsequently employed to guide the thought generation process. By leveraging this approach, L2T eliminates the need for task-specific prompts to direct the reasoning process. Instead, the labels effectively determine how the reasoning proceeds. In the following sections, we will elaborate on this process in detail. Figure \ref{fig:logic-graph} gives an illustration example for the reasoning process graph.

\subsection{Thought Generation Framework}

\begin{figure*}
    \centering
    \includegraphics[width=1\linewidth]{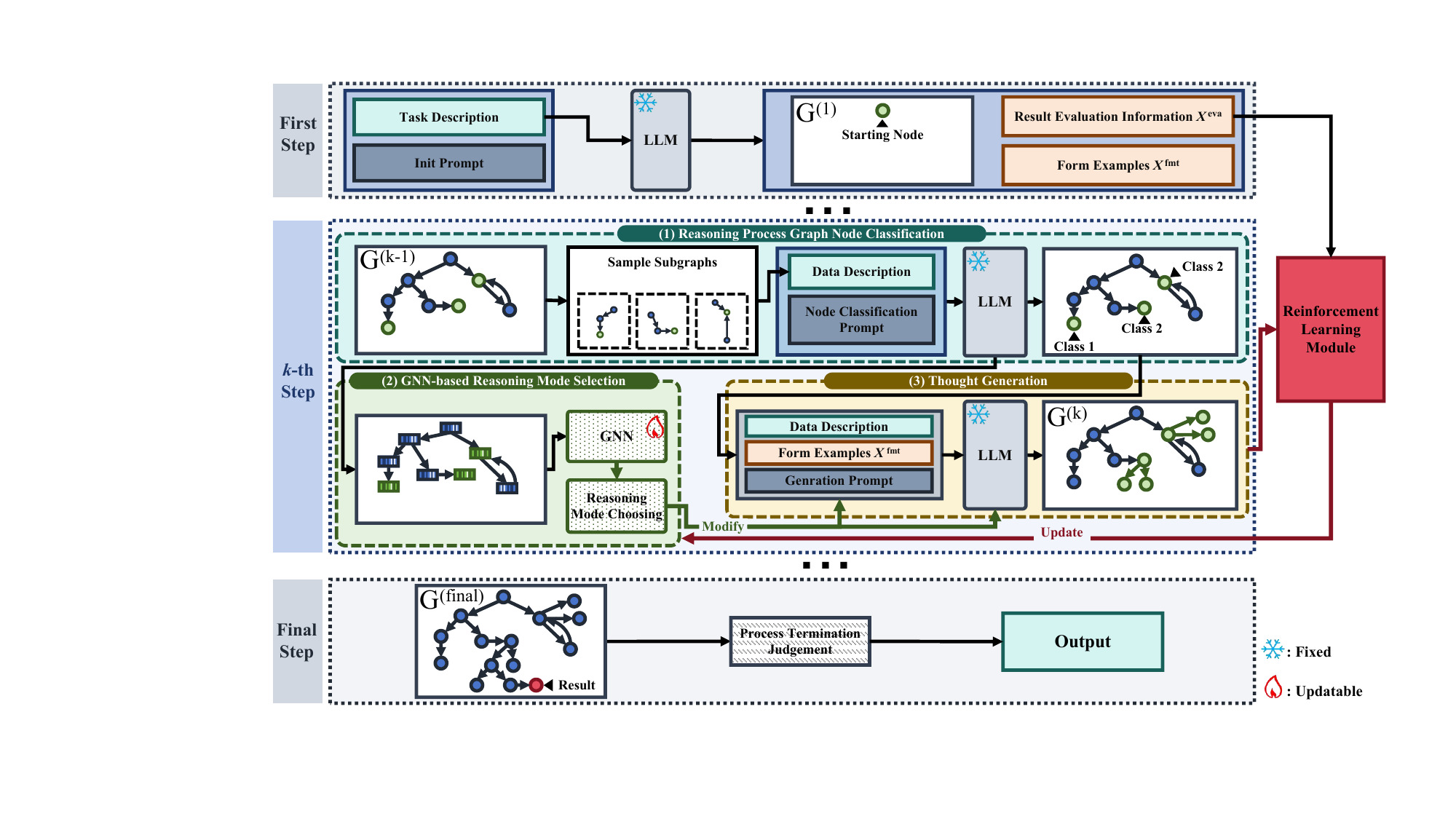}
    \vskip -0.05in
    \caption{The framework of the proposed method. All LLM modules uniformly utilize the same LLM.}
    % 最后更新该图??????
    \vskip -0.15in
    \label{fig:framework}
\end{figure*}

Next, we introduce our thought generation framework. Since the reasoning process is carried out step by step, we will explain in detail how reasoning is performed at the first step, the intermediate $k$-th step, and the final step, respectively. The overall framework is given in Figure \ref{fig:framework}.
% 使用了一个整体的框架，来进行reasoning process graph中各个待处理节点的分类，以基于类别来区分节点下一步的思维如何延展。之后使用GNN模块来引导自适应的模式切换

\subsubsection{First Step}
In the first step, we begin by obtaining an initial state. Using the LLM, we generate three components: the initial reasoning process graph $G^{(1)}$, the constraint format and examples for the process, and the evaluation criteria for the generated thoughts. Specifically, the initial reasoning process graph is defined as $G^{(1)} = \{\mathcal{V}^{(1)}, \mathcal{E}^{(1)}\}$. In \( G^{(1)} \), the subscript ``1'' in parentheses corresponds to the iteration step one. \( \mathcal{V}^{(1)} \) and \( \mathcal{E}^{(1)} \) represent the sets of nodes and edges in \( G^{(1)} \), respectively. At this stage, \( |\mathcal{V}^{(1)}| = 1 \) and \( \mathcal{E}^{(1)} = \varnothing \), as \( G^{(1)} \) contains only a single initial node. The node in \( G^{(1)} \) is assigned to  \( \mathcal{V}^{\text{pres} (1)} \). The attribute of this node is the task description.

Additionally, we utilize the LLM to directly produce \( X^{\text{fmt}} \) that includes format descriptions and a set of corresponding example answers for new thought generation. Furthermore, based on the LLM, we extract relevant information $X^{\text{eva}}$ from the task description that pertains to the criteria for evaluating and scoring the quality of the model's output. The above design ensures that L2T can perform step-by-step reasoning for complex problems in a unified format without relying on task-specific prompts, while also providing a reasonable evaluation of task execution. Details of all L2T prompts can be found in \textbf{Appendix \ref{apx:sgen}}.

\subsubsection{$k$-th Step}
For the $k$-th step, we generate the subsequent thoughts to construct $G^{(k)}$ based on $G^{(k-1)}$. L2T first conducts reasoning process graph node classification, then achieves GNN-based reasoning mode selection. Based on the classification information and the selected reasoning mode, L2T finally generates the thoughts. We will elaborate on the details in the following content.

\paragraph{(1) Reasoning process graph node classification.} The node classification is performed for all nodes within $\mathcal{V}^{\text{pres}(k-1)}$. For each node $v \in \mathcal{V}^{\text{pres} (k-1)}$, we extract its corresponding subgraph $\widetilde{G}_{v}^{(k-1)}$, where $\widetilde{G}_{v}^{(k-1)} = \{ \widetilde{\mathcal{V}}_{v}^{(k-1)}, \widetilde{\mathcal{E}}_{v}^{(k-1)} \}$. Here, $\widetilde{\mathcal{V}}_{v}^{(k-1)}$ represents the set of all nodes in $G^{(k-1)}$ that have paths pointing to node $v$ with a path length less than $\beta$, where $\beta$ is a predefined hyperparameter. The edge set $\widetilde{\mathcal{E}}_{v}^{(k-1)}$ is defined as:
\begin{align}
    \widetilde{\mathcal{E}}_{v}^{(k-1)} & = 
    \nonumber\\
    & \big\{(u, w) \in \mathcal{E}^{(k-1)} \mid u \in \widetilde{\mathcal{V}}_{v}^{(k-1)}, w \in \widetilde{\mathcal{V}}_{v}^{(k-1)}\big\}.
\end{align}
Thus, $\widetilde{G}_{v}^{(k-1)}$ is the induced subgraph of $G^{(k-1)}$ whose vertex set is $\widetilde{\mathcal{V}}_{v}^{(k-1)}$.
% 这些信息将被输入LLM，完成节点分类。除了v意外，其它节点与v的拓扑关系将会进行标注，由于相邻节点要么是历史信息，要么是回溯信息，所以该标准并不困难。我们在附录中给出了相应的具体Prompt和数据。（这里补充完整？？？？？？）
The provided information will be utilized as input to the LLM to perform node classification. Beyond the node attributes, the topological relationships between the target node \( v \) and other nodes will be annotated and expressed in textual form. This annotation process is straightforward, as the neighboring nodes primarily represent historical or backtracking information. The overall node classification process can be mathematically expressed as follows:
\begin{align}
    \dot{Y}_{v}^{(k)} & = f\left(S^{\text{node}}, \tau\left(\{x_{u} \mid u \in \widetilde{\mathcal{V}}_{v}^{(k-1)}\}, \widetilde{G}_{v}^{(k-1)}\right)\right),
\end{align}
% 体现一下Graph Representation Learning？？？？？？
where \( S^{\text{node}} \) represents the prompts designed for node classification, $\dot{Y}_{v}^{(k)}$ denotes the estimated label, \( x_{u} \) denotes the textual representation of the reasoning thought associated with node \( u \), $f(\cdot)$ denotes the LLM, and \( \tau(\cdot) \) is the function that converts graph-related information into a descriptive textual format. Further details regarding the implementation of $\tau(\cdot)$ are provided in \textbf{Appendix \ref{apx:tsn}}.

% 到此为止，我们基于LLM对于节点分类。接下来，我们将基于这些分类，有针对性的生成后续的思维。然而，正如在前面所讨论的那样，针对于不同的问题，不同的步骤，应该使用不同的推理模式，为了避免针对每个任务每个步骤进行微调，我们通过修改prompt结构以及LLM的超参数来做到这一点。具体来讲，我们基于GNN进行整个逻辑图的处理，并将其输出表示映射为向量$a$。每个\( \mathcal{V}^{\text{pres} (k)} \)中的思维节点$v^{(k)}$均对应一个向量$a^{(k)}_{v}$。$a^{(k)}_{v}$由一系列参数构成，其中包含了可调整prompt模式的参数，例如所生成分支数量，以及生成内容多少等，以及LLM的超参数，例如模型的温度参数和Top-p采样参数等。我们将$a^{(k)}_{v}$视为一种采取的动作，而将$G^{(k-1)}$视为状态，并采用强化学习范式对于GNN进行不断更新，以选择最优的模型，具体实现方法将在下一节中进行阐述。

\paragraph{(2) GNN-based reasoning mode selection.}
% Second, we conduct GNN-based reasoning mode selection. As discussed earlier, different problems and steps require distinct reasoning modes. To avoid fine-tuning for each task and step, we achieve this by modifying the prompts and adjusting the hyperparameters of the LLM. 
Our GNN-based reasoning mode selection module processes the graph $G^{(k-1)}$ using a GNN \cite{DBLP:conf/iclr/KipfW17,wu2020comprehensive} $g(\cdot)$, which is a deep learning model designed to process and analyze graph-structured data by leveraging the relationships between nodes and edges. The GNN takes an attributed graph as input and outputs feature vectors for each node. For the implementation of $g(\cdot)$, we utilize a one-layer Graph Convolutional Network (GCN) \cite{DBLP:conf/iclr/KipfW17} followed by a two-layer Multi-Layer Perceptron (MLP). During the aforementioned node classification, each node $v$ in $G^{(k-1)}$ save the final-layer representation $h_{v}$ generated by the LLM as the node feature vector for this stage. Here, $h_{v}$ is the representation corresponding to the last output token in the answer sequence. These output representations are subsequently transformed into vectors denoted as $\mathbf{a}$.
Specifically, each reasoning node \( v^{(k)} \) in \( \mathcal{V}^{\text{pres}(k)} \) is associated with a vector \(\mathbf{a}^{(k)}_{v}\). The vector \(\mathbf{a}^{(k)}_{v}\) consists of a set of parameters, including adjustable prompt-related parameters (e.g., the number of generated branches) as well as LLM hyperparameters (e.g., the temperature parameter). Formally, \(\mathbf{a}^{(k)}_{v}\) is defined as:
\begin{gather} 
    \mathbf{a}^{(k)}_{v} = A\big(g_{[v]}(G^{(k-1)})\big), 
\end{gather} 
where $A(\cdot)$ denotes the function that outputs $\mathbf{a}^{(k)}_{v}$ based on $g_{[v]}(G^{(k-1)})$, $g_{[v]}(G^{(k-1)})$ is the GNN output representation of node $v$ at the $(k-1)$-th step. In fact, $A(\cdot)$ implements the Actor mechanism in the Actor-Critic algorithm \cite{DBLP:conf/nips/KondaT99}, and the implementation details of this function will be elaborated in Section \ref{sec:update}. 
We treat \(\mathbf{a}^{(k)}_{v}\) as an action of choosing a mode of reasoning, the model will be iteratively updated to optimize the selection of modes. Further details regarding \(\mathbf{a}^{(k)}_{v}\) can be found in \textbf{Appendix \ref{apx:a}}.

\paragraph{(3) Thought generation.}
Finally, we carry out thought generation. According to the classes described in section \ref{sec:reasoning process graph}, for a given node \(v\), new nodes need to be generated only when the label of \(v\) is 2, and the newly generated nodes are all child nodes of \(v\). For other types of nodes, only deletion, modification, and adjustment of set membership are required, which can be directly addressed through standardized processing on \(G^{(k-1)}\). The standardized processing can be implemented through straightforward code development. 
Subsequently, we input the prompts, pre-generated template examples, and the textual description of node attributes into an LLM, enabling it to generate the subsequent thought nodes when the label of \(v\) is 2. The process of generating the textual features of a child node \(u\) based on the content of its parent node \(v\) can be formalized as follows:
\begin{gather}
    x_{u} = f(S^{\text{gen}}, x_{v}^{(k-1)}, X^{\text{fmt}}, \mathbf{a}^{(k)}_{v}),
\end{gather}
where \( x_{u} \) represents the textual features of the node \( u \), and \( S^{\text{gen}} \) denotes the prompt used for data generation. Note that a portion of the prompt is determined by \( \mathbf{a}^{(k)}_{v} \), which also influences the hyperparameters of the LLM. Based on \( x_{u} \), along with other standardized processing, the graph \( G^{(k)} \) can then be constructed. The newly generated child nodes, together with the backtracked parent nodes, will form $\mathcal{V}^{\text{pres}(k)}$.

\subsubsection{Final Step}
% 整个推理流程的结束以最终结果的出现作为结束。当判定当前的$\mathcal{V}^{\text{pres}}$中存在标签为3的节点，即判定出现最终结果的时候，整个流程结束。
The reasoning process concludes when the final result emerges. This occurs when the current set, denoted as \(\mathcal{V}^{\text{pres}}\), contains a node labeled as 3, signifying the appearance of the final result. At this point, all intermediate steps and iterations cease, and the process terminates.

Additionally, if all nodes in \(\mathcal{V}^{\text{pres}}\) have their corresponding thoughts labeled as 1 (indicating that reasoning stops at the current thought), these thoughts will then be regenerated. If they are still labeled as 1, the process will also terminate.

% 同时，如果V中的所有节点对应的thought都被标记为stop，则会将所有thought重新生成一边，如果还是标记为stop，则也终止流程。

\subsection{Update}
\label{sec:update}

We employ the Actor-Critic algorithm from reinforcement learning to optimize and update the GNN-based reasoning mode selection module, which comprises $g(\cdot)$ and $A(\cdot)$. These components work together to produce the output $\mathbf{a}_v^{(k)}$. The Actor-Critic algorithm uses two models: the Actor, which selects actions based on the policy, and the Critic, which evaluates the actions by estimating the value function to improve the policy. Please refer to \textbf{Appendix \ref{apx:ac}} and \textbf{\ref{apx:ppo}} for detailed introductions. Assuming we are at the $k$-th step, we first consider the case where there is only one node in $G^{(k-1)}$ that needs to be processed, i.e., $|\mathcal{V}^{\text{pres}(k)}|=1$, and the pending node is $v$. As mentioned in the previous section, at step $k$, we regard $\mathbf{a}^{(k)}_{v}$ as an action of choosing the mode of reasoning. At this point, $g_{[v]}(G^{(k-1)})$, i.e., the GNN output representation of node $v$ at the $(k-1)$-th step, is treated as the input state. 

The Actor, which is represented as $A(\cdot)$, is used to generate the action $\mathbf{a}_v^{(k)}$, which represents the selected reasoning mode. At the $k$-th step, we calculate an action distribution $\pi(\mathbf{a}_v^{(k)} | g_{[v]}(G^{(k-1)}); \theta_\text{actor})$ based on a single-layer MLP with $\theta_\text{actor}$ as the parameters, and action $\mathbf{a}_v^{(k)}$ is sampled from this distribution. The process can be formulated as:
\begin{gather}
\mathbf{a}_v^{(k)} \sim \pi(\mathbf{a}_v^{(k)} | g_{[v]}(G^{(k-1)}); \theta_\text{actor}),
\label{eq:td_error}
\end{gather}
$\pi$ denotes the strategy distribution. The parameters of $\pi$ is output with the MLP, which takes $g_{[v]}(G^{(k-1)})$ as its input. Next, we acquire an immediate reward $r_k$ and the next state $g_{[v]}(G^{(k)})$. The reward $r_k$ is set to 100 if the generated thought represents the final result. Otherwise, it is an integer between 0 and 10, determined by the LLM based on $G^{(k)}$ and $X^{\text{eva}}$. The detailed prompt used for this process is provided in \textbf{Appendix \ref{apx:sgen}}. 

The Critic evaluates the performance of the current strategy by estimating the state value function $V\big(g_{[v]}(G^{(k-1)})\big)$, which is also implemented using a single-layer MLP with $\theta_\text{critic}$ as the parameters.

We adopt the widely used PPO framework \cite{DBLP:journals/corr/SchulmanWDRK17} for LLM training as the specific implementation of the Actor-Critic algorithm, optimizing and updating the Actor and Critic that we have constructed. Through collaborative optimization, the policy network gradually learns a better strategy for selecting reasoning modes, enabling the model to dynamically optimize inference efficiency and performance under different graph states.

For graphs with multiple pending nodes, i.e., \( |\mathcal{V}^{\text{pres}(k)}| > 1 \), each node is processed sequentially as different steps, with optimization and updates performed individually.

\section{Experiments}

\begin{table*}[ht]\scriptsize
\renewcommand{\arraystretch}{1.3} % 行间距调整
\setlength{\tabcolsep}{6.8pt} % 列间距调整
\centering

\begin{tabular}{l|ccc|ccc|ccc|ccc}
\hline
\multirow{2}{*}{\textbf{Method}}  & \multicolumn{3}{c|}{\textbf{3$\times$3 Sudoku}} & \multicolumn{3}{c|}{\textbf{4$\times$4 Sudoku}} & \multicolumn{3}{c|}{\textbf{5$\times$5 Sudoku}} & \multicolumn{3}{c}{\textbf{4$\times$4 Sudoku w/o TSP}} \\ 
\cline{2-13}
& \textbf{Average} & \textbf{Min} & \textbf{Max} & \textbf{Average} & \textbf{Min} & \textbf{Max} & \textbf{Average} & \textbf{Min} & \textbf{Max} & \textbf{Average} & \textbf{Min} & \textbf{Max} \\
\hline
IO & 43.85$\pm$10.44 & 4/13 & 8/13 & 24.62$\pm$10.50 & 0/13 & 4/13 & 10.77$\pm$6.41 & 0/13 & 3/13 & \textit{ 24.62$\pm$10.50}  & \textit{0/13}  & \textit{4/13} \\
CoT (zero-shot) & 61.54$\pm$9.87 & 5/13 & 9/13  & 33.08$\pm$9.47 & 3/13 & 7/13 & 13.85$\pm$8.70 & 0/13 & 4/13 & 10.77$\pm$9.54 & 0/13 & 5/13 \\
CoT (few-shot) & 80.77$\pm$9.54 & 9/13 & 12/13 & 57.69$\pm$10.92 & 5/13 & 9/13 & 46.92$\pm$10.32 & 4/13 & 8/13 & 30.00$\pm$12.91 & 1/13 & 7/13 \\
ToT & 92.31$\pm$4.39 & 12/13 & 13/13 & 72.31$\pm$5.99 & 8/13 & 12/13 & 63.85$\pm$10.44 & 5/13 & 10/13 & 34.62$\pm$13.91 & 1/13 & 9/13 \\
GoT & 95.38$\pm$5.19 & 12/13 & 13/13 & 72.35$\pm$11.47 & 8/13 & 13/13 & 67.69$\pm$10.92 & 5/13 & 11/13 & 37.69$\pm$15.89 & 2/13 & 9/13 \\
AoT & 97.65$\pm$4.37 & \underline{12/13} & 13/13 & 77.69$\pm$7.25 & 8/13 & 12/13 & 69.41$\pm$9.66 & 8/13 & 12/13 & 36.47$\pm$13.67 & 2/13 & 9/13 \\
\hline
\textbf{L2T w/o GNN} & \underline{98.46$\pm$3.61} & 11/13 & \underline{13/13} & \underline{93.08$\pm$9.47} & \underline{9/13} & \underline{13/13} & \textbf{89.46$\pm$9.87} & 
\underline{9/13} & \textbf{13/13} & \textit{\underline{93.08$\pm$9.47}} & \textit{\underline{9/13}} & \textit{\underline{13/13}}  \\

\textbf{L2T} & \textbf{100.00$\pm$0.00} & \textbf{13/13} & \textbf{13/13} & \textbf{98.46$\pm$3.76} & \textbf{12/13} & \textbf{13/13} & \underline{89.23$\pm$6.41} & \textbf{10/13} & \underline{13/13} & \textit{\textbf{98.46$\pm$3.76}} & \textit{\textbf{12/13}} & \textit{\textbf{13/13}}  \\
\hline
\end{tabular}

\vskip -0.08in

\caption{Results for performance on Sudoku. \textbf{Bold} denotes the best result, and \underline{underline} denotes the second best. For tied results in either first or second place, the performance is determined by comparing other relevant results within the same group. Min and Max represent the best and worst performances achieved by a method, respectively, in terms of the number of correct solutions out of 13 total puzzle sets. Results for $4\times4$ Sudoku w/o TSP (without task-specific prompts) reflect the performance of models when task-specific prompts are removed. Since IO, L2T w/o GNN, and L2T do not use task-specific prompts by design, their results are directly copied from the corresponding problem and are shown in \textit{italic} to indicate this.}

\vskip -0.1in

\label{tab:sudoku}
\end{table*}

\begin{table}[ht]\scriptsize
\renewcommand{\arraystretch}{1.3} % 行间距调整
\setlength{\tabcolsep}{11pt} % 列间距调整
\centering
\begin{tabular}{l|cc}
\hline
\textbf{Method} & \textbf{Game of 24} & \textbf{Game of 24 w/o TSP} \\
\hline
IO & 15.92$\pm$1.89 & \textit{15.92$\pm$1.89} \\
CoT (zero-shot) & 28.63$\pm$0.86 & 25.82$\pm$2.01 \\
CoT (few-shot) & 30.34$\pm$2.21 & 26.12$\pm$2.23 \\
ToT & 70.52$\pm$3.26 & 48.12$\pm$1.18 \\
GoT & 72.30$\pm$1.55 & \textit{48.15$\pm$1.28} \\
AoT & 74.23$\pm$1.59 & \textit{27.54$\pm$7.76} \\
\hline
\textbf{L2T w/o GNN} & \underline{77.45$\pm$1.17} & \textit{\underline{77.45$\pm$1.17}} \\
\textbf{L2T} & \textbf{80.42$\pm$2.98} & \textit{\textbf{80.42$\pm$2.98}} \\
\hline
\end{tabular}

\vskip -0.08in

\caption{Results for performance on Game of 24. \textbf{Bold} denotes the best result, and \underline{underline} denotes the second best. Results for Game of 24 w/o TSP reflect the performance of models when task-specific prompts are removed. As Table \ref{tab:sudoku}, \textit{italic} denotes the results that are directly copied from the corresponding problem, as the corresponding method do not use task-specific prompts by design.}

\vskip -0.1in

\label{tab:24point}
\end{table}

\begin{table}[ht]\scriptsize
\renewcommand{\arraystretch}{1.3} % 行间距调整
\setlength{\tabcolsep}{2pt} % 列间距调整
\centering
\begin{tabular}{l|cccc}
\hline
\textbf{Method} & \textbf{3 Characters} & \textbf{4 Characters} & \textbf{5 Characters} & \textbf{3 Characters w/o TSP} \\
\hline
IO & 36.83$\pm$1.57 & 35.06$\pm$6.73 & 6.57$\pm$2.04 & \textit{36.83$\pm$1.57} \\
CoT (zero-shot) & 40.92$\pm$0.71 & 38.91$\pm$1.09 & 10.52$\pm$0.85 & 37.85$\pm$0.96 \\
CoT (few-shot) & 45.24$\pm$1.47 & 39.43$\pm$1.24 & 13.42$\pm$2.25 & 42.05$\pm$1.24 \\
ToT & 53.68$\pm$2.65 & 47.82$\pm$0.81 & 17.58$\pm$0.34 & 49.16$\pm$1.37 \\
GoT & 51.42$\pm$1.80 & 47.95$\pm$1.24 & 16.72$\pm$0.28 & 48.85$\pm$1.06 \\
AoT & 53.15$\pm$1.34 & 46.69$\pm$1.80 & 16.41$\pm$1.01 & 41.94$\pm$1.14 \\
\hline
\textbf{L2T w/o GNN} & \underline{67.74$\pm$1.09} & \underline{54.84$\pm$3.01} & \underline{25.81$\pm$2.58} & \textit{\underline{67.74$\pm$1.09}} \\
\textbf{L2T} & \textbf{69.31$\pm$0.64} & \textbf{59.75$\pm$0.99} & \textbf{27.93$\pm$0.05} & \textit{\textbf{69.31$\pm$0.64}} \\
\hline
\end{tabular}

\vskip -0.08in

\caption{Results for performance on TruthQuest. \textbf{Bold} denotes the best result, and \underline{underline} denotes the second best. Results for 3 Characters w/o TSP reflect the performance of models upon 3 Characters TruthQuest when task-specific prompts are removed. As Table \ref{tab:sudoku}, \textit{italic} denotes the results that are directly copied from the corresponding problem, as the corresponding method do not use task-specific prompts by design.}

\vskip -0.1in

\label{tab:TruthQuest}
\end{table}

\begin{table*}[ht]\scriptsize
\renewcommand{\arraystretch}{1.5} % 行间距调整
\setlength{\tabcolsep}{9pt} % 列间距调整
\centering
\begin{tabular}{l|cccc|cccc|cccc}
\hline
\multirow{2}{*}{\textbf{Method}} & \multicolumn{4}{c|}{\textbf{Sentence Formation (Less Hints)}} & \multicolumn{4}{c|}{\textbf{Sentence Formation (More Hints)}} & \multicolumn{4}{c}{\textbf{Text Expansion}} \\
\cline{2-13}
 & \textbf{Higher} & \textbf{Same} & \textbf{Lower} & \textbf{Std.}  & \textbf{Higher} & \textbf{Same} & \textbf{Lower} & \textbf{Std.}  & \textbf{Higher} & \textbf{Same} & \textbf{Lower} & \textbf{Std.}  \\
\hline
IO & 93.06 & 6.93 & 0.00 & $\pm$3.92  & 82.67 & 17.33 & 0.00 & $\pm$3.76 & 51.93 & 38.91 & 9.16 & $\pm$2.24 \\
CoT & 62.87 & 36.14 & 0.00 & $\pm$3.22 & 61.39 & 38.61 & 0.00 & $\pm$3.08 & 42.28 & 41.78 & 15.94 & $\pm$1.24 \\
ToT & 48.27 & 50.24 & 1.49 & $\pm$2.90 & 50.74 & 47.02 & 2.23 & $\pm$2.53 & 41.98 & 36.83 & 21.19 & $\pm$2.83 \\
GoT & 47.77 & 49.99 & 2.23 & $\pm$2.56 & 49.75 & 48.02 & 2.23 & $\pm$3.33 & 41.88 & 35.64 & 22.48 & $\pm$2.65 \\
AoT & 48.82 & 49.06 & 2.11 & $\pm$1.88 & 48.12 & 49.05 & 2.82 & $\pm$2.18 & 44.24 & 36.82 & 18.94 & $\pm$2.44 \\
\hline
\textbf{L2T w/o GNN}  & 15.05 & 50.84 & 34.11 & $\pm$3.38 & 15.38 & 39.13 & 45.48 & $\pm$3.84 & 15.88 & 64.08 & 20.04 & $\pm$4.10 \\
\hline
\end{tabular}

\vskip -0.05in

\caption{Comparison of method performance on the Creative Writing task. All data represent the performance of L2T comparisons to other methods. \textit{Higher} indicates cases where L2T achieved a better score compared to the corresponding method. \textit{Same} represents cases where L2T achieved the same score as the corresponding method. \textit{Lower} indicates cases where the L2T scored worse compared to the corresponding method.}

\vskip -0.1in

\label{tab:comparison}
\end{table*}

\begin{table}[ht]\scriptsize
\renewcommand{\arraystretch}{1.3} % 行间距调整
\setlength{\tabcolsep}{10pt} % 列间距调整
\centering
\begin{tabular}{l|c|c}
\hline
\textbf{Method} & \textbf{Accuracy (\%)} & \textbf{Generated Nodes} \\
\hline
L2T & 80.42$\pm$2.98 & 36.14$\pm$9.29 \\
L2T w MLP & 78.20$\pm$1.36 & 40.29$\pm$9.87 \\
L2T w/o RL & 78.85$\pm$1.42 & 43.13$\pm$8.61 \\
L2T w/o GNN & 77.45$\pm$1.17 & 46.56$\pm$21.11 \\
\hline
\end{tabular}

\vskip -0.1in

\caption{Comparison of accuracy and number of generated nodes for different methods.}

\vskip -0.1in

\label{tab:abl}
\end{table}

\begin{table}[ht]\scriptsize
\renewcommand{\arraystretch}{1.3} % 行间距调整
\setlength{\tabcolsep}{9pt} % 列间距调整
\centering
\begin{tabular}{l|ccc}
\hline
\multirow{2}{*}{\textbf{Method}} & \textbf{Prompt Tokens} & \textbf{Generate Tokens} & \multirow{2}{*}{\textbf{Tokens per Case}} \\
 & \textbf{per Thought} & \textbf{per Thought} &  \\
\hline
IO & 0.18k & 0.56k & 0.56k \\
CoT & 0.23k & 1.86k & 1.86k \\
AoT & 0.55k & 1.74k & 1.74k \\
ToT & 0.48k & 0.20k & 11.60k \\
GoT & 0.48k & 0.21k & 7.56k \\
L2T & 0.49k & 0.18k & 4.68k \\
\hline
\end{tabular}

\vskip -0.1in

\caption{Comparison of prompt tokens per thought, generate tokens per thought, and tokens per case for different methods.}

\vskip -0.1in

\label{tab:tokens}
\end{table}

% 在本节中，我们在多个不同的LLM推理任务上进行我们的方法的验证。在所有任务上，我们均使用了同样的prompt进行实验。

\subsection{Comparison with State-of-the-Art Methods}

\subsubsection{Baselines} 
For our experiments, we utilized GPT-4o as the base model. First, we directly compared our proposed L2T method with the original output of GPT-4o\cite{DBLP:journals/corr/abs-2303-08774} (denoted as IO). Subsequently, we compared L2T with several advanced LLM reasoning methods, including CoT \cite{DBLP:conf/nips/Wei0SBIXCLZ22}, ToT \cite{DBLP:conf/nips/YaoYZS00N23}, GoT \cite{DBLP:conf/aaai/BestaBKGPGGLNNH24}, and AoT \cite{DBLP:conf/icml/SelAK0024}. Among these, we specifically analyzed both the zero-shot and few-shot versions of CoT.

\subsubsection{Tasks}
% 我们统共采用了4种不同的任务，分别为数独，24点游戏，TruthQuest \cite{DBLP:conf/emnlp/MondorfP24}，创意性写作，均为同类方法常用的评价任务\cite{DBLP:conf/nips/YaoYZS00N23,DBLP:conf/aaai/BestaBKGPGGLNNH24}。数独任务是（这里用一句话简单介绍一下数独是啥），我们分别$3\times3$，$4\times4$，$5\times5$的数独上进行了实验。24点（这里用一句话简单介绍一下24点是啥）。TruthQuest\cite{DBLP:conf/emnlp/MondorfP24}则是一个新出现的LLM推理验证BenchMark。创意性写作任务则由一系列的写作任务（这里这个任务换个词，别和前面重复）组成，测试LLM的逻辑构思写作能力。在所有任务上，L2T均使用了同样的prompt进行实验。

We evaluated our method on four distinct tasks: Sudoku, the Game of 24, TruthQuest \cite{DBLP:conf/emnlp/MondorfP24}, and Creative Writing. These tasks were chosen as they are commonly used in the evaluation of similar methods \cite{DBLP:conf/nips/YaoYZS00N23,DBLP:conf/aaai/BestaBKGPGGLNNH24}.

The Sudoku task is a logic-based puzzle involving the placement of numbers within a grid according to specific rules, we adopted 3 sizes, $3\times3$, $4\times4$, and $5\times5$. Game of 24 is a mathematical puzzle where players use four given numbers and basic arithmetic operations to reach a total of 24. TruthQuest \cite{DBLP:conf/emnlp/MondorfP24} is a recently introduced benchmark for evaluating the reasoning and verification abilities of LLMs. Creative Writing task consisted of a series of diverse writing challenges (designed to avoid redundancy in task definitions) to assess the logical and conceptual abilities of LLMs in generating coherent and creative text. More details can be found in \textbf{Appendix \ref{apx:tasks}}.

For all tasks, the L2T method was tested using identical prompts, ensuring a consistent evaluation framework.

\subsubsection{Settings}
% 我们使用了ChatGPT-4o的API来进行所有的实验，包括baseline的，我们也分别测试了所有baseline在使用固定的prompt时候的表现，具体来讲，我们将CoT、ToT、GoT的任务特定部分去掉，完成了这部分的实验。其它setting细节可以在Appendix中找到。
We utilized the GPT-4o API to conduct all the experiments, including those for the baselines. We also present the performance of L2T w/o GNN, which refers to L2T without the GNN-based reasoning mode selection module and, as a result, does not require any training. 

Furthermore, we conducted additional experiments (marked in orange) that removed the task-specific components of methods including CoT, ToT, GoT, and AoT. Further details regarding the experimental settings and hyperparameter configurations can be found in \textbf{Appendix \ref{apx:Detailed Implementaiton}}.

\subsubsection{Results}
% 接下来我们将对于各个任务上的结果进行分析。表格1展示了在数独上的结果，表格2展示了在24点问题上的结果，表格3展示了在TruthQuest上的结果。可以看到，在各个问题上我们的方法均取得了最优的表现，同时相比其他方法有较大的提升。而且在不使用任务特定的提示的时候，我们的方法相对其它方法的优势就更加明显了，在该情况下，我们的方法十分显著地提升了LLM解决问题的效能。同时，L2T w/o GNN方法的表现也大大优于其它方法，这证明了即使不使用基于强化学习训练的GNN模块，我们的方法也可以取得更好的表现，证明即使在不对于我们的GNN进行轻量化训练的情况下，我们的方法也具有较大优势。

Next, we analyze the results across different tasks. Tables \ref{tab:sudoku}, \ref{tab:24point}, and \ref{tab:TruthQuest} summarize the results for Sudoku, Game of 24, and TruthQuest. Our method consistently outperforms others, showing significant improvements, particularly without task-specific prompts, where its efficiency advantage is more pronounced. Even without the GNN-based reasoning mode selection module (L2T w/o GNN), performance remains superior, highlighting the effectiveness of our approach.

Table \ref{tab:comparison} presents results on Creative Writing, focusing on relative scores to L2T. Evaluations via an LLM reduce fluctuations. L2T achieves higher or equivalent scores in over 80\% of cases, with less than 20\% lower, outperforming baselines. L2T w/o GNN performs comparably, supporting conclusions from prior results.

\subsection{In-Depth Analysis}
\subsubsection{Ablation Study}
% 为了实现对于算法的进一步深入分析, 我们又展开了消融实验。该实验在24点任务上对于我们的方法进行裁剪，以分析方法中各个模块的作用。我们实现了3种裁剪后的方法，其中，L2T w MLP为将方法中的GNN替换为MLP，L2T W/o RL 为不再使用强化学习方式对于GNN进行更新，而直接采用各个节点的评分直接对于GNN模块进行训练，L2T W/o GNN则为完全去掉GNN模块。 

% 实验结果如表\ref{tab:abl}所示。可以看到，我们不光对于各个方法的准确率进行了实验，并且也对于其所生成的节点数量进行了统计，这可以一定程度上反映算法得出最终结果所需要的思维步骤数量。从结果中可以看出，GNN模块对于L2T方法确实有着一定作用，但是其主要效能实际上却是表现在削减推理所需的步骤数量上。显然，加入了GNN模块的方法相比不加入的，所需的推理步骤数量较为明显的减少了。

To further delve into the analysis of our algorithm, we conducted ablation experiments. These experiments were performed on the Game of 24 task to evaluate the contribution of each component in our proposed method. We implemented three variations of the method with specific components ablated: (1) L2T w MLP, which replaces the GNN with an MLP; (2) L2T w/o RL, which removes the reinforcement learning mechanism for updating the GNN and instead directly trains the GNN-based reasoning mode selection module based on the scores of individual nodes; and (3) L2T w/o GNN, which completely eliminates the GNN-based reasoning mode selection module.

The experimental results are shown in Table \ref{tab:abl}. We not only evaluated the accuracy of each variant  but also analyzed the number of nodes generated by each method. This provides an indication of the number of reasoning steps required to arrive at the final result. The results demonstrate that the GNN-based reasoning mode selection module does contribute to the performance of the L2T method. However, its primary benefit lies in reducing the number of reasoning steps needed. Clearly, methods incorporating the GNN-based reasoning mode selection module require significantly fewer reasoning steps compared to those without it.

\subsubsection{Computational Consumption Analysis}
% 我们同时也对于算法的计算消耗进行了分析，我们使用token数量来衡量计算消耗。实验结果在表格\ref{tab:tokens}中展示。可以看到，L2T所耗费的计算资源与其他方法相比是持平的，同时优于GOT，这证明了L2T可以在不使用过多计算资源的情况下完成复杂推理任务并取得不错的效果。
We also analyzed the computational consumption of L2T, using the number of tokens as a metric to measure computational cost. The experimental results are presented in Table \ref{tab:tokens}. As shown, the computational resources consumed by L2T are comparable to those of other methods and outperform GoT. This demonstrates that L2T can accomplish complex reasoning tasks and achieve favorable results without requiring excessive computational resources. We also provide a detailed breakdown of the computational overhead in Table \ref{tab:acs}.

\begin{table}[h!]\scriptsize
\centering
\caption{Comparison of LLM access counts for different methods. \textbf{Bold} denotes the minimum value.}
\label{tab:acs}
\small
\begin{tabular}{l|c|c|c}
\hline
\textbf{Category} & \textbf{L2T} & \textbf{ToT} & \textbf{GoT} \\
\hline
24 Points & \textbf{26} & 48 & 30 \\

3 $\times$ 3 Sudoku & \textbf{22} & 32 & 28 \\

TruthQuest & \textbf{14} & 26 & 18 \\

Creative Writing & 21 & 32 & \textbf{20} \\
\hline
\end{tabular}
\end{table}

\begin{figure}[h]
    \centering
    % 第一张子图
    \subfigure[Results of Creative Writing.]{
        \includegraphics[width=0.45\linewidth]{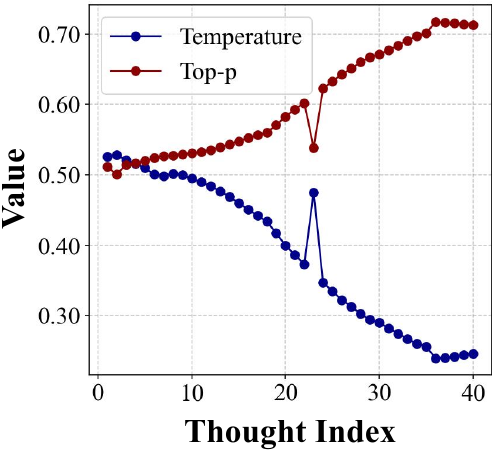}
        \label{fig:p1} % 标签
    }
    % 第二张子图
    \subfigure[Results of Game of 24.]{
        \includegraphics[width=0.45\linewidth]{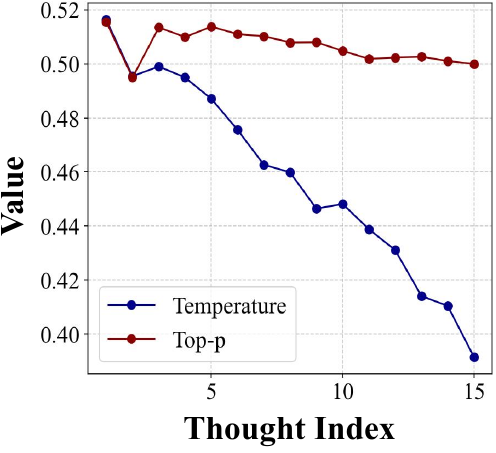}
        \label{fig:p2} % 标签
    }
        \subfigure[Visualization.]{
        \includegraphics[width=0.45\linewidth]{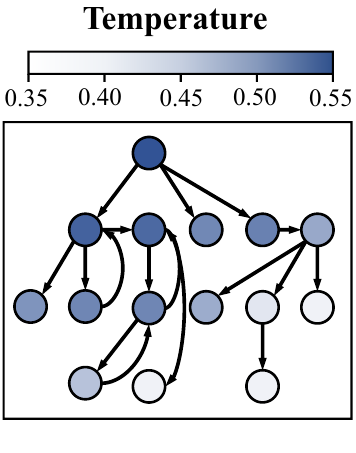}
        \label{fig:v} % 标签
    }

    % 总体标题
    \vskip -0.1in
    \caption{The temperature and top-\(p\) value within the reasoning process.} % 总标题
    \vskip -0.1in
    \label{fig:p}
\end{figure}

\subsubsection{Process Analysis}
% 为了对于L2T的工作过程进行更加深入的分析，我们将其运行过程中GNN模块输出的的Temperature和Top-p值进行了记录，其结果如图\ref{fig:p}所示。可以看到，一个有意思的现象是，Temperature和Top-p呈现明显的相关性。在Creative Writing上，该取值是一个大，另一个就小的（这里换成一个更加学术化的表述），而在Game of 24上，该取值又呈现出一个大，另一个也大的（这里换成一个更加学术化的表述）现象。这表明了，所训练的GNN模块针对不同任务采取了不同的策略，并且该策略关系是明显存在的。我们在图\ref{fig:v}中将该策略做了具象化的表示，以进行更加明显的展示。
In order to conduct a more in-depth analysis of the working process of L2T, we recorded the temperature and top-\(p\) values output by the GNN-based reasoning mode selection module during its operation. The results are shown in Figure \ref{fig:p}. An interesting observation is that temperature and top-\(p\) exhibit a significant correlation. For the Creative Writing task, the values display an inverse relationship—when one value is relatively high, the other tends to be relatively low. In contrast, for the Game of 24 task, the values show a direct relationship—when one value is high, the other is also high. This indicates that the trained GNN-based reasoning mode selection module adopts distinct strategies tailored to different tasks. To further clarify this, we provide a concrete visualization of this strategy in Figure \ref{fig:v}, offering a more explicit visualization of the parameter variations during the inference process is provided.

% \begin{figure}[h]
%     \centering
%     \includegraphics[width=0.5\linewidth]{graph.pdf}
%     \label{fig:motiv1} % 标签
%     % 总体标题
%     \vskip -0.1in
%     \caption{Visualization of the temperature value during reasoning process.} % 总标题
%     \vskip -0.1in
%     \label{fig:v}
% \end{figure}

\section{Conclusion}
% 本文提出了一种新的LLM推理方法，L2T。该方法采用图框架来对于LLM推理过程进行表示，同时采用图学习方法对于该推理过程图进行学习与分析，并且生成对应的推理策略。L2T采用了基于LLM的以及基于GNN的两种图学习方法，可以无需针对不同问题专门设计prompt，并且可以结合强化学习方法，在一次次处理问题的过程中不断自我优化。充足的实验证明了L2T的有效性。

% This paper proposes a novel LLM inference method, L2T. This method utilizes a graph-based framework to represent the inference process of LLMs and applies graph learning techniques to learn and analyze such a graph and subsequently generating corresponding inference strategies. Extensive experiments demonstrate the effectiveness of L2T.

This paper proposes a novel LLM reasoning method, L2T. This method utilizes a graph-based framework to represent the reasoning process of LLMs and applies graph learning techniques to learn and analyze this reasoning graph, subsequently generating corresponding reasoning strategies. L2T incorporates two types of graph learning approaches: one based on LLMs and the other based on GNNs. It eliminates the need for specifically designed prompts for different problems and can integrate reinforcement learning methods to continuously self-optimize during successive problem-solving processes. Extensive experiments demonstrate the effectiveness of L2T.

\section*{Acknowledgments}
We would like to express our sincere gratitude to the reviewers of this paper, as well as the Program Committee and Area Chairs, for their valuable comments and suggestions. This work is supported by the CAS Project for Young Scientists in Basic Research, Grant No. YSBR-040. 

\bibliographystyle{named}
\bibliography{ijcai25}

%%%%%%%%%%%%%%%%%%%%%%%%%%%%%%%%%%%%%%%%%%%%%%%%%%%%%%%%%%%%%%%%%%%%%%%%%%%%%%%
%%%%%%%%%%%%%%%%%%%%%%%%%%%%%%%%%%%%%%%%%%%%%%%%%%%%%%%%%%%%%%%%%%%%%%%%%%%%%%%
% APPENDIX
%%%%%%%%%%%%%%%%%%%%%%%%%%%%%%%%%%%%%%%%%%%%%%%%%%%%%%%%%%%%%%%%%%%%%%%%%%%%%%%
%%%%%%%%%%%%%%%%%%%%%%%%%%%%%%%%%%%%%%%%%%%%%%%%%%%%%%%%%%%%%%%%%%%%%%%%%%%%%%%

\newpage
\appendix
\onecolumn

\section{Implementation and Experimental Details}
\label{apx:Detailed Implementaiton}
\subsection{Implementation of the Actor-Critic Algorithm}
We used Proximal Policy Optimization (PPO) \cite{DBLP:journals/corr/SchulmanWDRK17} to implement the Actor-Critic algorithm \cite{DBLP:conf/nips/KondaT99}, leveraging its ability to stabilize policy optimization through constrained updates. PPO introduces a clipped surrogate objective that limits the magnitude of policy changes, ensuring stable training while maintaining the efficiency of policy gradient methods. 

\subsection{Hyperparameters}
The hyperparameters used during the experiments with the L2T model are as follows: the learning rate was set to $5 \times 10^{-3}$, reinforcement learning training was conducted over 20 epochs to refine decision-making strategies, the PPO clip parameter was set to 0.2 to regulate policy updates for stable learning, the maximum gradient norm was set to 0.5. Path hyperparameter $\beta$ was set to $2$.

\subsection{Implementation of Node Feature Extraction}
\label{apx:tsn}
The node feature extraction function $\tau(\cdot)$ is designed to extract the textual features of all nodes and organize them in a structured format. Specifically, the extracted content is formatted as: ``\textit{The former generated thoughts are: \{......\}, \{......\}, ......}'', where each individual thought is enclosed within curly brackets. This structured representation ensures that the thoughts generated from the nodes are clearly separated and easy to interpret. 

\subsection{Details Regarding Property Adjust Vector }
\label{apx:a}
% \(\mathbf{a}^{(k)}_{v}\)为一系列参数构成的向量，分为调节prompt的参数，以及调节LLM的参数，这些参数包括连续的（例如温度系数）以及离散的（例如分支数量）由GNN的输出中的某一维度直接投影或者经过softmax后进行类别判定。其中，调节prompt的参数使用了包括了分支数量、对于已经生成内容的依赖程度等。具体来讲，分支数量为节点后续生成的作为子节点的thought的数量，所提出的方法会根据\(\mathbf{a}^{(k)}_{v}\)中对应的数量来调整prompt，进而生成对应数量的thought。对于已经生成内容的依赖程度值将会直接嵌入prompt之中，提示后续生成内容做出相应调整。对于调节LLM的参数，则包括温度系数以及Top-p参数，以对于LLM进行具体调整。

\(\mathbf{a}^{(k)}_{v}\) represents a vector composed of a series of parameters, which can be categorized into two groups: parameters that adjust the prompt and parameters that fine-tune the behavior of the LLM. These parameters include both continuous values (e.g., temperature) and discrete values (e.g., the number of branches). These are either directly projected from a specific dimension of the output of the GNN or undergo a softmax operation for categorical determination.

Specifically, for the parameters that adjust the prompt, they include aspects such as the number of branches and the dependency on already generated content. The number of branches refers to the number of "thoughts" that are generated as child nodes for a given node. The proposed method adjusts the prompt according to the corresponding value in \(\mathbf{a}^{(k)}_{v}\), thereby generating the specified number of thoughts. The dependency on already generated content is directly embedded into the prompt to guide the subsequent generation, ensuring that it adapts to the context.

For parameters that fine-tune the behavior of the LLM, these include the temperature and top-\(p\) sampling parameters. Temperature adjusts the sharpness of the probability distribution over possible next tokens: lower temperatures make the model more deterministic, favoring high-probability tokens, while higher temperatures introduce more randomness and creativity. Top-$p$ sampling limits the candidate pool to the smallest set of tokens whose cumulative probability exceeds $p$, then samples from this set proportionally.

\subsection{Implementation of the Prompts}
\label{apx:sgen}

% 下面给出论文中的各个Prompts的实现方案。首先是生成thought 格式样例的prompt。
% I want to perform {task-specific content} task, and I want to break it down into steps. Please provide the specific content for each step, and how to design the input and output formats so that the task can be automated. Do not provide explanations, just give the answers directly.

The implementation details of various prompts used in the paper are provided below. First, we present the prompt for generating the format of a ``thought.''

\begin{tcolorbox}[colframe=black, colback=gray!20, coltitle=black, coltext=black, breakable,  boxrule=0.5mm, title=\textcolor{white}{Format Generation Prompt $X^{\text{fmt}}$}]
I aim to solve the task: \textbf{\textcolor{blue}{\textlangle\text{task description content}\textrangle}}. I want to solve it step by step. Please provide the specific content required to solve each step, along with the input and output formats, so that the task can be automated. Each solution must consist of at least two or more steps. Explanations are not needed; only clear and precise answers should be provided. Finally, please include three complete examples of the task execution process.
\end{tcolorbox}

% \textbf{\textcolor{blue}{$\textlangle\text{task description content}\textrangle$}} 为任务描述信息。所生成的格式与样例将被用于后续推理。推理过程中，将会提示模型自主选取对应的格式进行内容生成。该格式的作用主要用于统一形式，对于任务本身推理并无影响。请注意每次生成的格式不是固定的，这在后续结果中也有体现。
\textbf{\textcolor{blue}{\textlangle\text{task description content}\textrangle}} represents the task description information. The generated format and examples will be used for subsequent reasoning. During the reasoning process, the model will be prompted to autonomously select the corresponding format for content generation. The primary purpose of this format is to standardize the structure, and it does not directly influence the reasoning process for the task itself. It is important to note that the format generated each time is not fixed, which is also reflected in the subsequent results. The following prompt generates the evaluation critic information:
\begin{tcolorbox}[colframe=black, colback=gray!20, coltitle=black, coltext=black, breakable,  boxrule=0.5mm, title=\textcolor{white}{Evaluation Information Generation Prompt $X^{\text{eva}}$}]
I aim to solve the task: \textbf{\textcolor{blue}{\textlangle\text{task description content}\textrangle}}. I want to solve it step by step. Please provide the information related to the specific criteria required to assess each step. The evaluation criterion is to assess the degree of contribution of a specific step to the successful completion of the task, based on the task description. Please provide relevant information from the task.
\end{tcolorbox}
Then, we provide the evaluation prompt:
\begin{tcolorbox}[colframe=black, colback=gray!20, coltitle=black, coltext=black, breakable,  boxrule=0.5mm, title=\textcolor{white}{Evaluation Prompt }]
For the task: \textbf{\textcolor{blue}{\textlangle\text{task description content}\textrangle}}, \textbf{\textcolor{blue}{\textlangle\text{output results}\textrangle}} is a step in solving the task. Based on \textbf{\textcolor{orange}{\textlangle\text{evaluation information}\textrangle}}, evaluate whether this result is helpful in solving the task and rate it accordingly, outputting an integer from 0 to 10. Only output the integer.
\end{tcolorbox}

\textbf{\textcolor{blue}{\textlangle\text{output results}\textrangle}} denotes the generated thought, \textbf{\textcolor{orange}{\textlangle\text{evaluation information}\textrangle}} denotes the generated evaluating criteria $X^{\text{eva}}$. Next, we provide the implementation of prompt $S^{\text{node}}$.

\begin{tcolorbox}[colframe=black, colback=gray!20, coltitle=black, coltext=black, breakable,  boxrule=0.5mm, title=\textcolor{white}{Prompt $S^{\text{node}}$}]
% In order to solve task:\textbf{\textcolor{blue}{$\textlangle\text{task description content}\textrangle$}}, I need to solve it step by step. For each step, I generate a thought for problem-solving. 
To address the task: \textbf{\textcolor{blue}{\textlangle\text{task description content}\textrangle}}, I will break it down into step-by-step actions. For each step, I will generate a thought to solve the problem step by step. \textbf{\textcolor{blue}{\textlangle\text{related subgraph content}\textrangle}}, to generate the current thought, determine the appropriate action to take:

(1) Terminate: The current thought is incorrect and should be terminated. Provide a reason, but do not propose an alternative plan.

(2) Continues: Continue addressing the issue along the current line of thought.

(3) Complete: The problem has been successfully solved.

(4) Backtrack: The problem-solving process should be continued based on the former thought of current thought.

Answer by selecting the class of the action (from 1 to 4).
\end{tcolorbox}

\textbf{\textcolor{blue}{\textlangle\text{related subgraph content}\textrangle}} represents the generated related subgraph of thoughts in textual form. Then, we provide the implementation of prompt $S^{\text{gen}}$.

\begin{tcolorbox}[colframe=black, colback=gray!20, coltitle=black, coltext=black, breakable,  boxrule=0.5mm, title=\textcolor{white}{Prompt $S^{\text{gen}}$}]
To address the task: \textbf{\textcolor{blue}{\textlangle\text{task description content}\textrangle}}, I will decompose it into a series of step-by-step actions. For each step, I will generate a corresponding thought to systematically solve the problem. Considering the context provided in \textbf{\textcolor{blue}{\textlangle\text{related subgraph content}\textrangle}}, I will now proceed to address the issue by continuing along the current line of thought and generating the next step. The subsequent thought will be generated by selecting the appropriate step and following the specified format outlined in \textbf{\textcolor{orange}{\textlangle\text{format information}\textrangle}}. Generate \textbf{\textcolor{red}{\textlangle\text{branch number}\textrangle}} different thoughts.
\end{tcolorbox}
\textbf{\textcolor{orange}{\textlangle\text{format information}\textrangle}} denotes $X^{\text{fmt}}$. \textbf{\textcolor{red}{\textlangle\text{branch number}\textrangle}} denotes the number of the generated thoughts according to $\mathbf{a}^{(k)}_{v}$. 

% \begin{tcolorbox}[colframe=black, colback=gray!20, coltitle=black, coltext=black, breakable,  boxrule=0.5mm, title=\textcolor{white}{Type Generation Prompt}]
% \textbf{Given data:}

% The following contents are the descriptions of nodes within a graph: \textbf{\textcolor{blue}{$\textlangle\text{node attribute 1}\textrangle$}}; \textbf{\textcolor{blue}{$\textlangle\text{node attribute 2}\textrangle$}}; \textbf{\textcolor{blue}{$\textlangle\text{node attribute 3}\textrangle$}}; \textcolor{blue}{......}; \textbf{\textcolor{blue}{$\textlangle\text{node attribute n}\!>$}}.

% \textbf{Answer the following questions:}
% \begin{enumerate}
%     \item Which \textbf{\textcolor{orange}{$\textlangle\text{format type number}\textrangle$}} types can these nodes be divided according to their format? Provide the names of these types and separate them with semicolons.
    
%     \item Which \textbf{\textcolor{red}{$\textlangle\text{content type number}\textrangle$}} types can these nodes be divided according to their content? Provide the names of these types and separate them with semicolons.
% \end{enumerate}
% \textbf{Please only provide answers and separators strictly in the given order.}
% \end{tcolorbox}

\section{Tasks}
\label{apx:tasks}
\subsection{Game of 24}
Game of 24 is a mathematical reasoning challenge, where the goal is to use 4 numbers and basic
 arithmetic operations ($+-\times\div$) to obtain 24. We utilize the same
dataset proposed in \cite{DBLP:conf/nips/YaoYZS00N23}, which has 1,362 games that are sorted from easy to hard by human solving time, and use a subset of relatively hard games indexed 901-1,000 for testing.
\subsection{Sudoku Puzzles}
The Sudoku puzzles involve filling the numbers from $1$ to $n$ in an $n \times n$ grid, ensuring that each row and each column contains no repeated numbers. We use the benchmark proposed in \cite{long2023large}, which contains $3 \times 3$,$4 \times 4$ and $5 \times 5$ Sudoku Puzzles.
\subsection{TruthQuest}
We use the TruthQuest benchmark proposed in \cite{DBLP:conf/emnlp/MondorfP24},we use part of the dataset, and test on 3,4 and 5 characters. TruthQuest are Knights and Knaves puzzles, a class of
logical puzzles in which each character is either a
"Knight" or a "Knave". The fundamental rule of
these puzzles is that a Knight always tells the truth,
meaning that any statement made by a Knight is
logically consistent with the facts. In contrast, a
Knave always lies, meaning that every statement
made by a Knave is false. The objective of an LLM
agent is to logically deduce the identity of each
character based on their statements.
\subsection{Creative Writing}
We propose two different creative writing tasks.The first task is to provide several words, expand each word into a sentence containing it, and then combine the sentences into a complete paragraph.
The second task is to provide four short sentences, expand each into a small paragraph starting with the sentence, and then combine them into a complete paragraph.The words and short sentences were all randomly generated by the LLM.

\section{Reasoning Processes}

% 在本节中，我们给出推理的具体过程中LLM的输出，来更好的对于我们的方法进行展示
In this section, we provide the specific outputs of the reasoning process generated by the LLM to better demonstrate our approach. \textbf{Generated Thought Content} denotes the output thought, \textbf{Label} denotes the class of the thought.

\subsection{Game of 24}
% \hrule
% \vspace{0.2cm} % 增加线和文字之间的距离
\colorbox{gray!20}{\parbox{\linewidth}{\textbf{Based on Thought 0 (root) to generate Thought 1,2,3,4, and 5:}}}\\
\\
\textbf{Generated Thought Content:}
Input:[10,9,2,3] Plan:10 + 2 = 12 Output:[9,3,12]\\
\textbf{Label:} 2 (Reasoning Contiue)\\
\\
\textbf{Generated Thought Content:}
Input:[10,9,2,3] Plan:9 - 3 = 6 Output:[10,2,6]\\
\textbf{Label:} 2 (Reasoning Contiue)\\
\\
\textbf{Generated Thought Content:}
Input:[10,9,2,3] Plan:10 - 9 = 1 Output:[1,2,3]\\
\textbf{Label:} 2 (Reasoning Continue)\\
\\
\textbf{Generated Thought Content:}
Input:[10,9,2,3] Plan:3 * 2 = 6 Output:[10,9,6]\\
\textbf{Label:} 2 (Reasoning Continue)\\
\\
\textbf{Generated Thought Content:}
Input:[10,9,2,3] Plan:9 + 3 = 12 Output:[10,2,12]\\
\textbf{Label:} 2 (Reasoning Continue)\\
\\
\colorbox{gray!20}{\parbox{\linewidth}{\textbf{Based on Thought 1 to generate Thought 6,7,8,9, and 10:}}}\\
\\
\textbf{Generated Thought Content:}
Input:[9,3,12] Plan:12 - 9 = 3 Output:[3,3]\\
\textbf{Label:} 1 (Reasoning Stop)\\
\\
\textbf{Generated Thought Content:}
Input:[9,3,12] Plan:12 / 3 = 4 Output:[9,4]\\
\textbf{Label:} 2 (Reasoning Continue)\\
\\
\textbf{Generated Thought Content:}
Input:[9,3,12] Plan:9 + 3 = 12 Output:[12,12]\\
\textbf{Label:} 2 (Reasoning Continue)\\
\\
\textbf{Generated Thought Content:}
Input:[9,3,12] Plan:9 * 3 = 27 Output:[27,12]\\
\textbf{Label:} 1 (Reasoning Stop)\\
\\
\textbf{Generated Thought Content:}
Input:[9,3,12] Plan:12 - 3 = 9 Output:[9,9]\\
\textbf{Label:} 2 (Reasoning Continue)\\
\\
\colorbox{gray!20}{\parbox{\linewidth}{\textbf{Based on Thought 2 to generate Thought 11,12,13,14, and 15:}}}\\
\\
\textbf{Generated Thought Content:}
Input:[10,2,6] Plan:10 + 2 = 12 Output:[12,6]\\
\textbf{Label:} 2 (Reasoning Continue)\\
\\
\textbf{Generated Thought Content:}
Input:[10,2,6] Plan:10 - 6 = 4 Output:[4,2]\\
\textbf{Label:} 2 (Reasoning Continue)\\
\\
\textbf{Generated Thought Content:}
Input:[10,2,6] Plan:10 * 2 = 20 Output:[20,6]\\
\textbf{Label:} 2 (Reasoning Continue)\\
\\
\textbf{Generated Thought Content:}
Input:[10,2,6] Plan:6 / 2 = 3 Output:[10,3]\\
\textbf{Label:} 2 (Reasoning Continue)\\
\\
\textbf{Generated Thought Content:}
Input:[10,2,6] Plan:10 / 2 = 5 Output:[5,6]\\
\textbf{Label:} 2 (Reasoning Continue)\\
\\
\colorbox{gray!20}{\parbox{\linewidth}{\textbf{Based on Thought 3 to generate Thought 16,17,18,19, and 20:}}}\\
\\
\textbf{Generated Thought Content:}
Input:[1,2,3] Plan:1 + 2 = 3 Output:[3,3]\\
\textbf{Label:} 1 (Reasoning Stop)\\
\\
\textbf{Generated Thought Content:}
Input:[1,2,3] Plan:2 + 3 = 5 Output:[1,5]\\
\textbf{Label:} 1 (Reasoning Stop)\\
\\
\textbf{Generated Thought Content:}
Input:[1,2,3] Plan:3 - 1 = 2 Output:[2,2]\\
\textbf{Label:} 1 (Reasoning Stop)\\
\\
\textbf{Generated Thought Content:}
Input:[1,2,3] Plan:3 * 2 = 6 Output:[1,6]\\
\textbf{Label:} 2 (Reasoning Continue)\\
\\
\textbf{Generated Thought Content:}
Input:[1,2,3] Plan:3 / 1 = 3 Output:[2,3]\\
\textbf{Label:} 1 (Reasoning Stop)\\
\colorbox{gray!20}{\parbox{\linewidth}{\textbf{Based on Thought 4 to generate Thought 21,22,23, and 24:}}}\\
\\
\textbf{Generated Thought Content:}
Input:[10,9,6] Plan:10 + 9 = 19 Output:[19,6]\\
\textbf{Label:} 1 (Reasoning Stop)\\
\\
\textbf{Generated Thought Content:}
Input:[10,9,6] Plan:10 - 6 = 4 Output:[9,4]\\
\textbf{Label:} 2 (Reasoning Continue)\\
\\
\textbf{Generated Thought Content:}
Input:[10,9,6] Plan:10 * 6 = 60 Output:[9,60]\\
\textbf{Label:} 1 (Reasoning Stop)\\
\\
\textbf{Generated Thought Content:}
Input:[10,9,6] Plan:9 + 6 = 15 Output:[10,15]\\
\textbf{Label:} 2 (Reasoning Continue)\\
\\
\colorbox{gray!20}{\parbox{\linewidth}{\textbf{Based on Thought 5 to generate Thought 25,26,27, and 28:}}}\\
\\
\textbf{Generated Thought Content:}
Input:[10,2,12] Plan:10 + 2 = 12 Output:[12,12]\\
\textbf{Label:} 1 (Reasoning Stop)\\
\\
\textbf{Generated Thought Content:}
Input:[10,2,12] Plan:12 - 10 = 2 Output:[2,2]\\
\textbf{Label:} 1 (Reasoning Stop)\\
\\
\textbf{Generated Thought Content:}
Input:[10,2,12] Plan:12 / 2 = 6 Output:[10,6]\\
\textbf{Label:} 1 (Reasoning Stop)\\
\\
\textbf{Generated Thought Content:}
Input:[10,2,12] Plan:12 - 2 = 10 Output:[10,10]\\
\textbf{Label:} 1 (Reasoning Stop)\\
\\
\colorbox{gray!20}{\parbox{\linewidth}{\textbf{Based on Thought 7 to generate Thought 29,30,31,32, and 33:}}}\\
\\
\textbf{Generated Thought Content:}
Input:[9,4] Plan:9 - 4 = 5 Output:[5]\\
\textbf{Label:} 2 (Reasoning Continue)\\
\\
\textbf{Generated Thought Content:}
Input:[9,4] Plan:9 / 4 = 2.25 Output:[2.25]\\
\textbf{Label:} 1 (Reasoning Stop)\\
\\
\textbf{Generated Thought Content:}
Input:[9,4] Plan:9 + 4 = 13 Output:[13]\\
\textbf{Label:} 1 (Reasoning Stop)\\
\\
\textbf{Generated Thought Content:}
Input:[9,4] Plan:9 * 4 = 36 Output:[36]\\
\textbf{Label:} 1 (Reasoning Stop)\\
\\
\textbf{Generated Thought Content:}
Input:[9,4] Plan:4 - 9 = -5 Output:[-5]\\
\textbf{Label:} 1 (Reasoning Stop)\\
\\
\colorbox{gray!20}{\parbox{\linewidth}{\textbf{Based on Thought 8 to generate Thought 34,35,36, and 37:}}}\\
\\
\textbf{Generated Thought Content:}
Input:[12,12] Plan:12 + 12 = 24 Output:[24]\\
\textbf{Label:} 3 (Reasoning Finished)\\
\\
\textbf{Generated Thought Content:}
Input:[12,12] Plan:12 + 12 = 24 Output:[24]\\
\\
\textbf{Generated Thought Content:}
Input:[12,12] Plan:12 / 2 = 6 Output:[6]\\
\\
\textbf{Generated Thought Content:}
Input:[12,12] Plan:12 - 12 = 0 Output:[0]\\
\\

\subsection{Creative Writing}
\colorbox{gray!20}{\parbox{\linewidth}{\textbf{Based on Thought 0(root) to generate Thought 1 and 2:}}}\\
\\
\textbf{Generated Thought Content:}\\
Input:["Elephant","Solar","Lantern","Velvet"]\\
Plan:choose element 0 and element 2\\
Output:["The elephant gently lifted the lantern with its trunk, illuminating the path ahead.","Solar","Velvet"]\\
\textbf{Label:} 2 (Reasoning Continue)\\
\\
\textbf{Generated Thought Content:}\\
Input:["Elephant","Solar","Lantern","Velvet"]\\
Plan:choose element 1 and element 3\\
Output:["The solar panels absorbed the sunlight, their surface smooth like velvet.","Elephant","Lantern"]\\
\textbf{Label:} 2 (Reasoning Continue)\\
\\
\colorbox{gray!20}{\parbox{\linewidth}{\textbf{Based on Thought 1 to generate Thought 3, 4, and 5:}}}\\
\\
\textbf{Generated Thought Content:}\\
Input:["The elephant gently lifted the lantern with its trunk, illuminating the path ahead.","Solar","Velvet"]\\
Plan:choose element 1 and element 2\\
Output:["The solar panels absorbed the sunlight, their surface smooth like velvet.","The elephant gently lifted the lantern with its trunk, illuminating the path ahead."]\\
\textbf{Label:} 2 (Reasoning Continue)\\
\\
\textbf{Generated Thought Content:}\\
Input:["The elephant gently lifted the lantern with its trunk, illuminating the path ahead.","Solar","Velvet"]\\
Plan:choose element 0 and element 1\\
Output:["The elephant gently lifted the lantern with its trunk, illuminating the path ahead. The solar panels absorbed the sunlight, powering the lantern.","Velvet"]\\
\textbf{Label:} 2 (Reasoning Continue)\\
\\
\textbf{Generated Thought Content:}\\
Input:["The elephant gently lifted the lantern with its trunk, illuminating the path ahead.","Solar","Velvet"]\\
Plan:choose element 0 and element 2\\
Output:["The elephant gently lifted the lantern with its trunk, illuminating the path ahead. The soft velvet of the cushion on the bench invited her to sit.","Solar"]\\
\textbf{Label:} 2 (Reasoning Continue)\\
\\
\colorbox{gray!20}{\parbox{\linewidth}{\textbf{Based on Thought 2 to generate Thought 6 and 7:}}}\\
\\
\textbf{Generated Thought Content:}\\
Input:["The solar panels absorbed the sunlight, their surface smooth like velvet.","Elephant","Lantern"]\\
Plan:choose element 1 and element 2\\
Output:["The elephant gently lifted the lantern with its trunk, illuminating the path ahead.","The solar panels absorbed the sunlight, their surface smooth like velvet."]\\
\textbf{Label:} 2 (Reasoning Continue)\\
\\
\textbf{Generated Thought Content:}\\
Input:["The solar panels absorbed the sunlight, their surface smooth like velvet.","Elephant","Lantern"]\\
Plan:choose element 0 and element 1\\
Output:["The solar panels absorbed the sunlight, their surface smooth like velvet. The elephant stood nearby, its massive form casting a shadow.","Lantern"]\\
\textbf{Label:} 2 (Reasoning Continue)\\
\\
\colorbox{gray!20}{\parbox{\linewidth}{\textbf{Based on Thought 3 to generate Thought 8:}}}\\
\\
\textbf{Generated Thought Content:}\\
Input:["The solar panels absorbed the sunlight, their surface smooth like velvet.","The elephant gently lifted the lantern with its trunk, illuminating the path ahead."]\\
Plan:choose element 0 and element 1\\
Output:["The solar panels absorbed the sunlight, their surface smooth like velvet. The elephant gently lifted the lantern with its trunk, illuminating the path ahead."]\\
\textbf{Label:} 2 (Reasoning Continue)\\
\\
\colorbox{gray!20}{\parbox{\linewidth}{\textbf{Based on Thought 4 to generate Thought 9:}}}\\
\\
\textbf{Generated Thought Content:}\\
Input:["The elephant gently lifted the lantern with its trunk, illuminating the path ahead. The solar panels absorbed the sunlight, powering the lantern.","Velvet"]\\
Plan:choose element 0 and element 1\\
Output:["The elephant gently lifted the lantern with its trunk, illuminating the path ahead. The solar panels absorbed the sunlight, powering the lantern. The soft velvet of the cushion on the bench invited her to sit."]\\
\textbf{Label:} 2 (Reasoning Continue)\\
\\
\colorbox{gray!20}{\parbox{\linewidth}{\textbf{Based on Thought 5 to generate Thought 12 and 13:}}}\\
\\
\textbf{Generated Thought Content:}\\
Input:["The elephant gently lifted the lantern with its trunk, illuminating the path ahead. The soft velvet of the cushion on the bench invited her to sit.","Solar"]\\
Plan:choose element 0 and element 1\\
Output:["The elephant gently lifted the lantern with its trunk, illuminating the path ahead. The soft velvet of the cushion on the bench invited her to sit. The solar panels absorbed the sunlight, powering the lantern."]\\
\textbf{Label:} 2 (Reasoning Continue)\\
\\
\textbf{Generated Thought Content:}\\
Input:["The elephant gently lifted the lantern with its trunk, illuminating the path ahead. The soft velvet of the cushion on the bench invited her to sit.","Solar"]\\
Plan:choose element 1 and element 0\\
Output:["The solar panels absorbed the sunlight, powering the lantern. The elephant gently lifted the lantern with its trunk, illuminating the path ahead. The soft velvet of the cushion on the bench invited her to sit."]\\
\textbf{Label:} 2 (Reasoning Continue)\\
\\
\colorbox{gray!20}{\parbox{\linewidth}{\textbf{Based on Thought 6 to generate Thought 14 and 15:}}}\\
\\
\textbf{Generated Thought Content:}\\
Input:["The elephant gently lifted the lantern with its trunk, illuminating the path ahead.","The solar panels absorbed the sunlight, their surface smooth like velvet."]\\
Plan:choose element 0 and element 1\\
Output:["The elephant gently lifted the lantern with its trunk, illuminating the path ahead. The solar panels absorbed the sunlight, their surface smooth like velvet."]\\
\textbf{Label:} 2 (Reasoning Continue)\\
\\
\textbf{Generated Thought Content:}\\
Input:["The elephant gently lifted the lantern with its trunk, illuminating the path ahead.","The solar panels absorbed the sunlight, their surface smooth like velvet."]\\
Plan:choose element 1 and element 0\\
Output:["The solar panels absorbed the sunlight, their surface smooth like velvet. The elephant gently lifted the lantern with its trunk, illuminating the path ahead."]\\
\textbf{Label:} 2 (Reasoning Continue)\\
\\
\colorbox{gray!20}{\parbox{\linewidth}{\textbf{Based on Thought 8 to generate Thought 14 and 15:}}}\\
\\
\textbf{Generated Thought Content:}\\
Input:["The solar panels absorbed the sunlight, their surface smooth like velvet. The elephant stood nearby, its massive form casting a shadow.","Lantern"]\\
Plan:choose element 0 and element 1\\
Output:["The solar panels absorbed the sunlight, their surface smooth like velvet. The elephant stood nearby, its massive form casting a shadow. The lantern's glow softly reflected off the panels, creating a warm ambiance."]\\
\textbf{Label:} 2 (Reasoning Continue)\\
\\
\textbf{Generated Thought Content:}\\
Input:["The solar panels absorbed the sunlight, their surface smooth like velvet. The elephant stood nearby, its massive form casting a shadow.","Lantern"]\\
Plan:choose element 1 and element 0\\
Output:["The lantern's glow softly reflected off the solar panels, their surface smooth like velvet. The elephant stood nearby, its massive form casting a shadow."]\\
\textbf{Label:} 2 (Reasoning Continue)\\
\\
\section{Further Backgrounds}

\subsection{Graph Neural Networks}
Graph Neural Networks (GNNs) \cite{DBLP:conf/iclr/KipfW17,wu2020comprehensive} are a class of neural networks designed to process graph-structured data. The key idea is to update each node's representation by aggregating information from its neighbors, allowing the model to learn graph-level or node-level representations. A graph \( G = (\mathcal{V}, \mathcal{E}) \) consists of a set of nodes \( \mathcal{V} \) and edges \( \mathcal{E} \). Each node \( v \in \mathcal{V} \) has an associated feature vector \( \mathbf{x}_v \in \mathbb{R}^d \), and an edge \( (u, v) \in \mathcal{E} \) indicates a relationship between nodes \( u \) and \( v \). In GNNs, the representation of each node is updated by aggregating information from its neighbors. The message passing process consists of two main steps. First, for each node \( v \), information from its neighbors \( N(v) \) is aggregated. The most common aggregation operations are sum, mean, or max pooling. The update for node \( v \) is given by:
\begin{gather}
\mathbf{h}_v^{(k)} = \text{AGGREGATE}\left( \left\{ \mathbf{h}_u^{(k-1)} : u \in N(v) \right\} \right),
\end{gather}
where \( \mathbf{h}_v^{(k)} \) is the representation of node \( v \) at the \( k \)-th layer, and \( \mathbf{h}_u^{(k-1)} \) is the representation of node \( u \) at the previous layer. Second, the aggregated information is passed through a neural network (usually an MLP) to update the node representation. The update rule for node \( v \) is:
\begin{gather}
\mathbf{h}_v^{(k)} = \sigma\left( W^{(k)} \cdot \left( \mathbf{h}_v^{(k-1)} \oplus \mathbf{h}_v^{(k)} \right) + b^{(k)} \right),
\end{gather}
where \( W^{(k)} \) and \( b^{(k)} \) are the weight matrix and bias for the \( k \)-th layer, \( \oplus \) denotes concatenation of node feature vector and aggregated neighbor information, and \( \sigma \) is the activation function. After several iterations of message passing, each node's representation captures more information from its neighbors. For graph-level tasks, such as graph classification, the entire graph's representation can be obtained by pooling the node representations, which is done by:
\begin{gather}
\mathbf{h}_G = \text{POOL}\left( \left\{ \mathbf{h}_v^{(K)} : v \in \mathcal{V} \right\} \right),
\end{gather}
where \( \mathbf{h}_G \) is the graph representation and \( K \) is the number of message passing layers. During training, GNNs typically use supervision based on graph or node labels. For node classification, the loss function is commonly the cross-entropy loss, represented by:
\begin{gather}
\mathcal{L} = - \sum_{v \in \mathcal{V}} y_v \log(\hat{y}_v),
\end{gather}
where \( y_v \) is the true label for node \( v \), and \( \hat{y}_v \) is the predicted label for node \( v \).

\subsection{Actor-Critic Algorithm}
\label{apx:ac}
The Actor-Critic algorithm \cite{DBLP:conf/nips/KondaT99} combines policy gradient methods (Actor) and value estimation methods (Critic). The main steps are as follows:

\paragraph{1. Initialization.}
Initialize the parameters of the policy network (Actor) and the value network (Critic), typically with random initialization, and initialize the environment and state.

\paragraph{2. Interaction with the Environment.}
At each time step, the agent selects an action \( a_t \) based on the policy output from the Actor:
\begin{equation}
a_t = \pi_\theta(s_t),
\end{equation}
where \( \pi_\theta(s_t) \) represents the probability distribution over actions \( a_t \) given state \( s_t \), and \( \theta \) is the parameter of the Actor.

\paragraph{3. Execute Action and Observe Results.}
After executing action \( a_t \), the environment returns the next state \( s_{t+1} \) and reward \( r_t \).

\paragraph{4. Update Critic.}
The Critic evaluates the goodness of the action by computing the state value function \( V(s_t) \). The Critic is updated using the Temporal Difference (TD) error:
\begin{equation}
\delta_t = r_t + \gamma V(s_{t+1}) - V(s_t),
\end{equation}
where \( \gamma \) is the discount factor, and \( \delta_t \) is the TD error. The Critic's parameters \( \theta_{\text{critic}} \) are updated as follows:
\begin{equation}
\theta_{\text{critic}} \leftarrow \theta_{\text{critic}} + \alpha_{\text{critic}} \delta_t \nabla_{\theta_{\text{critic}}} V(s_t),
\end{equation}
where \( \alpha_{\text{critic}} \) is the learning rate of the Critic.

\paragraph{5. Update Actor.}
The Actor updates the policy by optimizing the objective function. Typically, policy gradient methods are used, and the Critic's value estimate is used to update the policy. The goal of the Actor is to maximize the expected reward, and the update rule is:
\begin{equation}
\theta_{\text{actor}} \leftarrow \theta_{\text{actor}} + \alpha_{\text{actor}} \delta_t \nabla_{\theta_{\text{actor}}} \log \pi_\theta(s_t, a_t),
\end{equation}
where \( \alpha_{\text{actor}} \) is the learning rate of the Actor, \( \delta_t \) is the TD error, and \( \log \pi_\theta(s_t, a_t) \) is the log probability of selecting action \( a_t \) in state \( s_t \).

\paragraph{6. Repeat Steps.}
Repeat steps 2 to 5 until a stopping condition is met (e.g., reaching the maximum number of training steps or convergence).

\subsection{PPO Algorithm}
\label{apx:ppo}
PPO \cite{DBLP:journals/corr/SchulmanWDRK17} improves upon the Actor-Critic framework by introducing a "proximal optimization" strategy to ensure the stability of each policy update. The main improvements of PPO are as follows:

\paragraph{1. Clipped Importance Sampling.}
PPO introduces a clipping mechanism to limit the magnitude of each update, preventing overly large policy updates. Specifically, PPO uses an importance ratio \( r_t(\theta) \) to measure the ratio between the new and old policies, and clips this ratio:
\begin{equation}
L^{\text{CLIP}}(\theta) = \hat{\mathbb{E}}_t \left[ \min \left( r_t(\theta) \hat{A}_t, \text{clip}(r_t(\theta), 1 - \epsilon, 1 + \epsilon) \hat{A}_t \right) \right],
\end{equation}
where \( r_t(\theta) = \frac{\pi_\theta(a_t|s_t)}{\pi_{\theta_{\text{old}}}(a_t|s_t)} \) is the importance ratio, \( \hat{A}_t \) is the advantage estimate, and \( \epsilon \) is the clipping threshold.

\paragraph{2. Advantage Estimation.}
PPO uses the advantage function \( \hat{A}_t \) to measure how good a particular action is relative to the current policy. The advantage function is typically computed using Generalized Advantage Estimation (GAE):
\begin{equation}
\hat{A}_t = \delta_t + (\gamma \lambda) \delta_{t+1} + \cdots + (\gamma \lambda)^{T-t+1} \delta_T,
\end{equation}
where \( \delta_t = r_t + \gamma V(s_{t+1}) - V(s_t) \) is the TD error, \( \gamma \) is the discount factor, and \( \lambda \) is the GAE parameter.

\paragraph{3. Objective Function Optimization.}
The objective function in PPO is based on the Actor-Critic algorithm and incorporates the clipping strategy:
\begin{equation}
L^{\text{PPO}}(\theta) = \hat{\mathbb{E}}_t \left[ \min \left( r_t(\theta) \hat{A}_t, \text{clip}(r_t(\theta), 1 - \epsilon, 1 + \epsilon) \hat{A}_t \right) \right],
\end{equation}
ensuring stable training by limiting the magnitude of the policy update.

\paragraph{4. Multiple Epochs of Updates.}
PPO performs multiple updates (usually 3-4) on each sampled batch during training to improve sample efficiency and accelerate convergence.

\end{document}